\documentclass[10pt,twocolumn,letterpaper]{article}

\usepackage{wacv}              

\usepackage{multirow} 
\usepackage{algorithm}
\usepackage{algpseudocode}

\definecolor{wacvblue}{rgb}{0.21,0.49,0.74}
\usepackage[pagebackref,breaklinks,colorlinks,allcolors=wacvblue]{hyperref}

\def\wacvPaperID{324} 
\def\confName{WACV}
\def\confYear{2026}

\usepackage{xspace}
\newcommand\ours{\texttt{MuSACo}\xspace}
\newcommand\oursuda{\texttt{MuSACo (UDA)}\xspace}
\newcommand\oursmsda{\texttt{MuSACo (MSDA)}\xspace}

\title{\ours: \underline{Mu}ltimodal \underline{S}ubject-Specific Selection and \underline{A}daptation for Expression Recognition with \underline{Co}-Training}

\author{\parbox{16cm}{\centering
    {\large Muhammad Osama Zeeshan$^1$, Natacha Gillet$^2$, Alessandro Lameiras Koerich$^1$, \\Marco Pedersoli$^1$, Francois Bremond$^3$, Eric Granger$^1$}\\ 
    {\normalsize
    $^1$  LIVIA, Dept. of Systems Engineering, ÉTS Montreal, Canada\\
    $^2$ École Polytechnique, Palaiseau, France
    \\
    $^3$ Centre INRIA d’Université Côte d’Azur, Sophia Antipolis, France} \\
     {\normalsize muhammad-osama.zeeshan.1@ens.etsmtl.ca,
    \{eric.granger, marco.pedersoli, alessandro.koerich\}@etsmtl.ca}\\
    {\normalsize natacha.gillet@polytechnique.edu, francois.bremond@inria.fr}
}}

\begin{document}
\maketitle

\begin{abstract}
Personalized expression recognition (ER) involves adapting a machine learning model to subject-specific data for improved recognition of expressions with considerable interpersonal variability. Subject-specific ER can benefit significantly from multi-source domain adaptation (MSDA) methods -- where each domain corresponds to a specific subject -- to improve model accuracy and robustness. Despite promising results, state-of-the-art MSDA approaches often overlook multimodal information or blend sources into a single domain, limiting subject diversity and failing to explicitly capture unique subject-specific characteristics. To address these limitations, we introduce \ours, a multimodal subject-specific selection and adaptation method for ER based on co-training. It leverages complementary information across multiple modalities and multiple source domains for subject-specific adaptation. This makes \ours particularly relevant for affective computing applications in digital health, such as patient-specific assessment for stress or pain, where subject-level nuances are crucial.
\ours selects source subjects relevant to the target and generates pseudo-labels using the dominant modality for class-aware learning, in conjunction with a class-agnostic loss to learn from less confident target samples. Finally, source features from each modality are aligned, while only confident target features are combined. Experimental results
on challenging multimodal ER datasets -- BioVid, StressID, and BAH -- show that \ours outperforms UDA (blending) and state-of-the-art MSDA methods. Our code is available: \url{https://github.com/osamazeeshan/MuSACo}
\end{abstract}
\vspace{-10pt}

\section{Introduction}
\label{sec:intro}

\begin{figure*}
\centering
\includegraphics[width=1.0\linewidth]{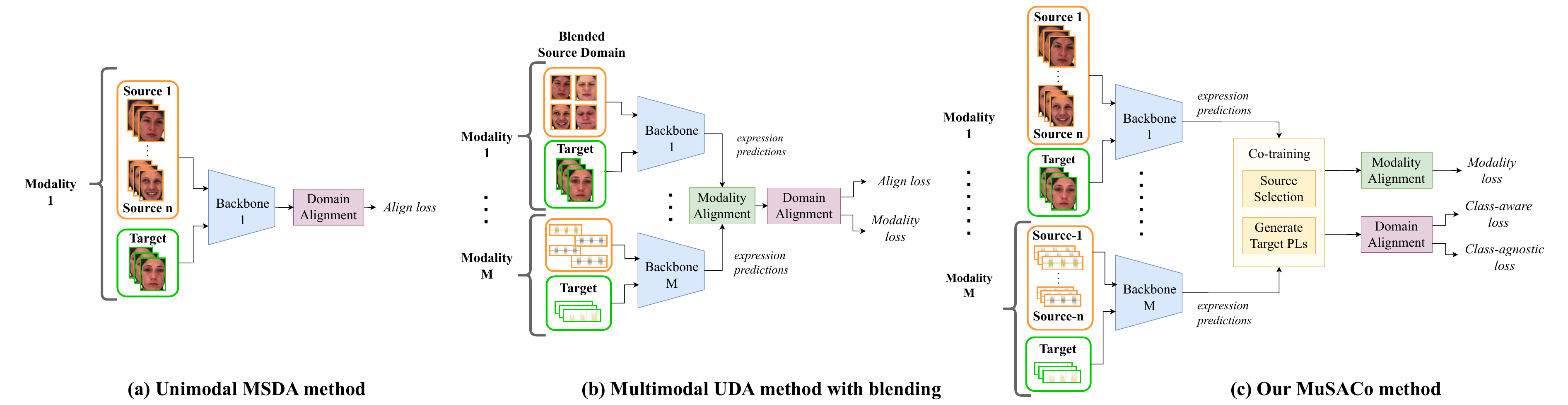}
\caption{Comparison of \ours against unimodal MSDA and multimodal UDA (blending) ER methods for subject-based adaptation. \textbf{(a) Unimodal MSDA} aligns multiple source subjects to the target within a single modality, which reduces its accuracy. \textbf{(b) Multimodal UDA with blending} incorporates single or blended source domains for target adaptation but does not fully exploit multiple subject-specific diversity. \textbf{(c) \ours} selects relevant sources per modality using co-training, and aligns them to the target using both class-aware and class-agnostic losses before fusing modalities for final prediction.}
\label{fig:mm_motiv}
\vspace{-10pt}
\end{figure*}

Expression recognition (ER) has achieved significant success in computer vision due to its relevance in numerous real-world applications, such as pain estimation, stress monitoring, and affective computing \cite{aslam2023privileged, aslam2024distilling, li2018deep, waligora2024joint, sharafi2025disentangled}. These models can achieve a high level of performance under supervised settings or when provided with limited labeled data \cite{praveen2020deep, rajasekhar2021deep}. However, their effectiveness mainly depends on the availability of annotated datasets, which are difficult to obtain, and often fail to generalize for subtle expressions and across diverse individuals due to variations in cultural, environmental, and individual expressiveness~\cite{jack2009cultural}. Unsupervised domain adaptation (UDA)  \cite{gretton2012kernel, long2016unsupervised, xu2018deep, guichemerre2024source} has emerged as a promising direction to address this challenge by adapting a model from a labeled source domain to an unlabeled target domain. In the context of ER, this facilitates personalization, where a model trained on labeled expression data from multiple source subjects is adapted to a new, unlabeled target subject.
While existing UDA approaches \cite{han2020personalized, li2018deep, zhu2016discriminative} adapt to the unlabeled target dataset treated as a domain by either using single or blended labeled source subjects, recent advances in multi-source domain adaptation (MSDA) \cite{zeeshan2024subject, zeeshan2025progressive} leverages information from multiple source subjects (domains). This encourages diversity in the source model for adaptation to an unlabeled target subject (domain). This method facilitates more precise and targeted adaptation strategies for developing a personalized ER model (see Fig.~\ref{fig:mm_motiv}(a)).

Despite these advances, the majority of subject-specific ER approaches are predominantly constrained to a single modality~\cite{han2020personalized, zeeshan2025progressive}. 
However, human expressions and fine-grained affective cues are often difficult to capture reliably from a single modality alone \cite{liu2021comparing, zhang2020emotion}. To address this limitation, we focus on a multimodal setting that leverages the complementary strengths of multimodal data.
Multimodal ER (MER) has been widely studied in the literature \cite{aslam2024distilling, aslam2023privileged, praveen2022joint, tzirakis2017end}, typically outperforming unimodal systems by leveraging complementary information from multiple modalities. MER approaches integrate diversity from various modalities, such as visual, textual, audio, or physiological signals \cite{dzedzickis2020human} to enhance model performance. In subject-based ER, most MER methods focus on fully or weakly labeled data \cite{tzirakis2017end, praveen2022joint}, hindering the ability of the model to have a better adaptation to unlabeled data. Nevertheless, there are some studies \cite{jimenez2024cfda, wang2022multi} that introduced multiple modalities, specifically for EEG. Eye movement signals with a single source or combine multiple sources into a single domain (see Fig.~\ref{fig:mm_motiv}(b)). Their reliance on a single source often fails to provide diversity in source domains. Furthermore, blending multiple sources into a single domain \cite{jimenez2024mmda} cannot explicitly capture the unique subject-specific characteristics of each source subject, which is crucial for subject-specific target adaptation.
In this paper, we propose \ours, a novel approach for \underline{M}ultimodal \underline{S}ubject-Specific Selection and \underline{A}daptation for ER using \underline{Co}-Training in an MSDA setting (Fig.~\ref{fig:mm_motiv}(c)). Our method exploits the complementary information from multiple modalities while selecting the most relevant source subjects for the target. It takes advantage of the unique characteristics of each subject for better performance on the target subject (domain\footnote{The terms domain and subject are used interchangeably.}). There are two key challenges when considering multiple modalities for subject-specific MSDA: (1) effectively leveraging multiple source subject information within multimodal systems, and (2) achieving cross-domain and cross-modal alignment.

To address the first challenge, we propose the \emph{selection of source subjects} most relevant to the target subject. While prior work~\cite{zeeshan2024subject} has shown that selecting source subjects based on their relevance to the target improves adaptation, it relies solely on a single modality to guide the source selection process. In this paper, we propose a co-training-based strategy for selecting source subjects that leverage multiple modalities, as different modalities complement each other by providing unique characteristics from various perspectives. Thus, selecting sources using co-training benefits the adaptation process. Furthermore, to improve generalization across individuals, we encourage the model to focus on expression-specific features by disentangling identity-related information during source training.

For the second challenge, we first focus on cross-domain alignment. State-of-the-art MSDA methods \cite{kang2020contrastive, scalbert2021multi} highlight the effectiveness of class-aware alignment in bridging dissimilar distributions. This typically involves generating pseudo-labels (PLs) for the target subject. In our proposed method, we extend this by leveraging co-training for the \emph{generation of target PLs}. Specifically, PLs for unlabeled target data are generated by selecting predictions from the dominant modality (e.g., visual or physiological), chosen by their probability score and threshold. This ensures diversity in modality-specific feature representations and captures complementary aspects of the data. These PLs then guide class-aware alignment between source and target subjects, minimizing distribution mismatch. Although PLs allow improving the model performance, they heavily depend on the choice of a threshold to filter out unreliable samples. In particular, high threshold values ensure the sample quality. While setting a high threshold value often results in discarding potentially useful but less confident samples. To mitigate this, we introduce a class-agnostic loss that aligns non-confident target samples with the source. This encourages the unreliable samples to contribute to the learning process to improve the robustness of the model. For modality alignment, including a fusion module \cite{praveen2022joint, aslam2023privileged} has proven to enhance MER performance. In \ours, a fusion module is added to concatenate features from different modalities for each selected source subject. For the target, only reliable samples (selected via co-training) are combined.

The contributions of this paper are summarized as follows. 
\textbf{(1)} \ours: a novel multimodal MSDA method for accurate subject-specific adaptation. It selects relevant source subjects from the target using co-training, leveraging complementary information across modalities.
\textbf{(2)} An effective alignment process that relies on co-training to generate confident target PLs for class-aware loss, combined with class-agnostic loss to handle non-confident target samples. 
\textbf{(3)} Extensive experiments performed on challenging MER datasets (BioVid, StressID, and BAH) show the efficacy of \ours. It can outperform the UDA (blending) and state-of-the-art MSDA methods on this multimodal data.

\section{Related Work}
\label{sec:relatedwork}

\noindent \textbf{(a) Multi-Source Domain Adaptation: }
MSDA techniques for image classification can be classified into {discrepancy-based} methods \cite{kang2020contrastive, peng2019moment} that reduce the domain shift by estimating the discrepancy between domains. Contrastive-based approaches \cite{kang2020contrastive, scalbert2021multi} generate target pseudo-labels using class-wise clustering, while adversarial and self-supervised methods \cite{zhao2021madan, nguyen2021stem, venkat2020your, deng2022robust} train multiple source-specific classifiers and derive target pseudo-labels through classifier agreement for domain adaptation. Recently, \citet{zeeshan2024subject} proposed a subject-based MSDA method that treats each subject as a domain to handle many sources. However, existing MSDA approaches remain unimodal (visual-only), motivating our introduction of multimodality for subject-based ER.

\noindent \textbf{(b) Multimodal Expression Recognition: }
Multimodal ER has been widely studied. Cross-attention fusion has been proposed to combine multiple modalities for expression recognition \cite{praveen2022joint}, while LUPI-based methods \cite{aslam2023privileged} and optimal transport distillation \cite{aslam2024distilling} enable learning from multiple modalities but inference with fewer. In domain adaptation, multimodal approaches have been applied to EEG and eye-tracking, e.g., variational autoencoders with cycle consistency \cite{wang2022multi} or CORAL-based alignment with intra/inter-domain losses \cite{jimenez2024cfda}. Building on these, our work leverages visual and physiological modalities in a multi-source DA setting to adapt to unlabeled target subjects.

\noindent \textbf{(c) Disentanglement in Expression Recognition: }
Disentanglement methods suppress identity or pose factors that hinder ER. GAN-based models such as \cite{jiang2022disentangling, xie2020facial} rely on image reconstruction, making them effective for visual data but unsuitable for non-visual modalities like physiology. In contrast, KNIFE \cite{pichler2022differential} disentangles by minimizing entropy between estimators, a modality-agnostic approach. Motivated by this, \ours adopts a similar formulation to decouple identity from expression representations.   

\noindent \textbf{(d) Co-training for Domain Adaptation: }
Co-training is a well-established semi-supervised and unsupervised learning approach \cite{blum1998combining} that is used in many applications, such as semantic segmentation \cite{xia2020uncertainty, gomez2023co}, task decomposition \cite{yang2021deep}, or in image classification \cite{chen2011co}. It typically trains the model with modalities to produce more reliable pseudo-labels (PLs), while domain adaptation usually trains the model with a single modality that incorporates multiple classifiers to account for different views. Chen et al. \cite{chen2011co} propose a method that introduces two different classifiers trained with different features controlled via a weighting mechanism while generating target pseudo-labels from the most confident classifier. \ours introduces a co-training method for selecting source subjects from multimodal data. In addition, we select target confident PLs based on multiple classifiers trained on different modalities.  

\section{Proposed \ours Approach}

\begin{figure*}[t!]
\centering
\includegraphics[width=1.0\linewidth]{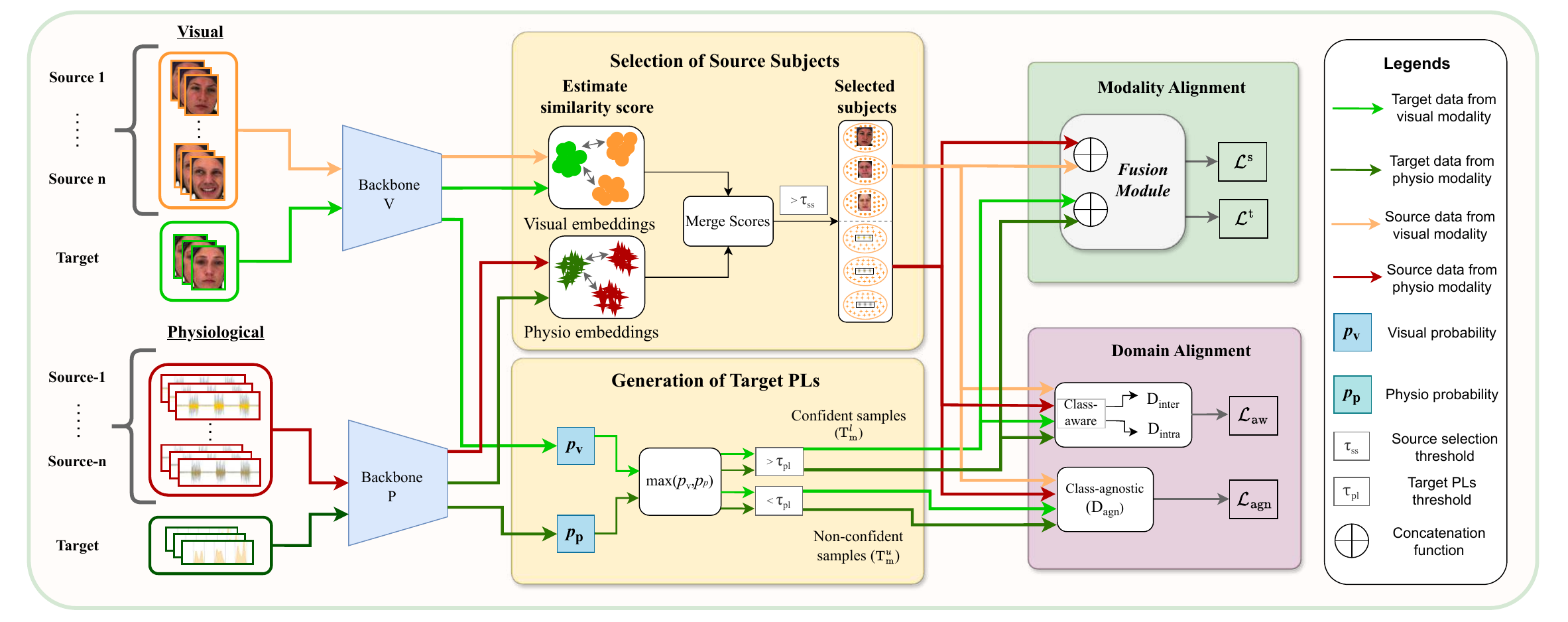}
\caption{An overview of \ours shown in the particular case of $M$ = 2 modalities (with visual + physiological). First, the similarity between source and target subjects is estimated using the \textbf{Selection of Source Subjects} module, followed by selecting the modality that gives the maximum probability score based on the threshold. Then, \textbf{Domain Alignment} is achieved by the \textbf{Generation of Target PLs} module through co-training. It calculates class-aware loss for each modality, combined with class-agnostic loss to learn from the non-confident target samples. Finally, the \textbf{Fusion Module} is introduced for the \textbf{Modality Alignment} through feature concatenation.}
\label{fig:dimsa_network}      
\vspace{-10pt}
\end{figure*}

\ours seeks to leverage complementary information from multiple modalities within MSDA using co-training to improve target personalization. This is accomplished through a two-step training process. Firstly, the non-discriminative identity information is disentangled from the source subjects while training task-specific backbones using source supervision. This ensures that only discriminative features are retained for recognizing the expression. Second, a co-training strategy is employed to select the most relevant source subjects for the target. For each source-target pair, similarity scores are computed from each modality. Sources that produce high similarity scores in either modality are selected, and a threshold $(\tau_{ss})$ is applied to filter out less relevant sources. Exploiting the trained backbones, we generate target PLs through co-training. The modality with the highest confidence in its prediction is chosen, and the samples are considered reliable if their confidence exceeds a threshold $(\tau_{pl})$. For domain alignment, class-aware alignment is performed on the target samples selected using $\tau_{pl}$ and class-agnostic alignment using non-confident samples. Finally, for modality alignment, features are concatenated from different modalities for every selected source and target subject, which are used to train the fusion module. \ours is illustrated in Fig.\ref{fig:dimsa_network}.

\subsection{Model Architecture}

A multimodal multi-source expression classification task is considered, where a model is trained using several labeled source subjects $\mathcal{S}_m={\{\mathbf{S}_{m_1},\ldots,\mathbf{S}_{m_a},\ldots,\mathbf{S}_{m_D} \}}$ and a single unlabeled target subject $\mathbf{T}_m$, where $m$ is the number of modalities and $a={\{1,\dots,D\}}$ is the number of source subjects. In \ours, without loss of generality, we consider two modalities: \textbf{visual} ($\text{v}$) and \textbf{physiological} ($\text{p}$), where $m$ $\in{\{\text{v},\text{p}\}}$.
Each source $\mathbf{S}_{m_a}$ contains $N^{\textbf{s}}$ labeled samples ${\{(\mathbf{x}_{m_j}^\textbf{s}, y_{m_j}^{\textbf{s}}, \acute{y}_{m_j}^{\textbf{s}})|1<j<{N^\textbf{s}}}\}$, where $y_{m_j}^{\textbf{s}}$ represents expression labels and $\acute{y}_{m_j}^{\textbf{s}}$ represents identity labels. We also define expression classes as ${c=\{1,\dots,C\}}$ 
The unlabeled target consists of $N^\textbf{t}$ samples ${\{\mathbf{x}_{m_j}^\textbf{t}|1<j<{N^\textbf{t}}\}}$. \textbf{Backbones} are defined as $\mathcal{B}={\{B_1,\dots,B_M\}}$, where each $B$ belongs to its respective modality. Accordingly, we consider two backbones that correspond to $B_\text{v}$ (visual) and $B_\text{p}$ (physiological) modalities. It takes input from $\mathbf{x}_{m}^\textbf{s}$ and $\mathbf{x}_{m}^\textbf{t}$ and return embeddings of source $\mathbf{h}^\textbf{s}_m=\mathbf{h}_\text{v}^\textbf{s}, \mathbf{h}_\text{p}^\mathbf{s}$ and target $\mathbf{h}^\textbf{t}_m=\mathbf{h}_\text{v}^\textbf{t}, \mathbf{h}_\text{p}^\textbf{t}$. 
Finally, the \textbf{fusion module} $F$ is a two-layer perceptron that takes input extracted from  $\mathcal{S}_m$ and $\mathbf{T}_m$ to perform feature fusion for the alignment of modalities.

\subsection{Training Backbones with Disentanglement}
\label{sec:meth_train_bb}
Two backbones are trained for each modality -- visual $B_\text{v}$ and physiological $B_\text{p}$ -- using multiple labeled source subjects $\mathcal{S}_m$. To learn expression-discriminative representations, each modality is supervised using a cross-entropy:
\begin{equation}
    \mathcal{L}_{m}^{\textbf{s}}=\frac{1}{D}\frac{1}{N^{\textbf{s}}}\sum_{i=1}^{D}\sum_{j=1}^{N^{\textbf{s}}}\text{CE}(\mathbf{y}_{m_j}^i,f_{m}(\mathbf{h}_{m_j}^i))
\end{equation}
where $\mathbf{h}_{m_j}^i$ is the embedding from modality $m$, $\mathbf{y}_{m_j}^i$ is the ground-truth expression label, and $f_m$ is the modality-specific expression classifier. The loss $\mathcal{L}_m^{\textbf{s}}$ is computed independently for each modality to ensure effective learning of modality-specific features.

\noindent\textbf{Disentangled Identity-Information.}
Our approach is inspired by KNIFE~\cite {pichler2022differential}, a differentiable entropy estimator, to disentangle identity information from expression-relevant features in each modality. We estimate the marginal entropy $H(\mathbf{h}_m^\mathbf{s})$ and the conditional entropy $H(\mathbf{h}_{m_j}^\mathbf{s}|\acute{Y}_{m_j}^\mathbf{s})$. For each sample in the source subjects, the identity labels $\acute{y}_{m_{j}}^\textbf{s}$ are converted into one-hot representations $\acute{Y}_{m_j}^\textbf{s} \in R^{k}$ and define the disentanglement loss as:
\begin{equation}
\mathcal{L}_{m}^d=H(\mathbf{h}_m^\textbf{s})+H(\mathbf{h}_{m_j}^\textbf{s}|\acute{Y}_{m_j}^\textbf{s}).
\end{equation} 
The classification loss $\mathcal{L}_m^s$ and disentanglement loss $\mathcal{L}_m^d$ are jointly optimized to preserve task-relevant features while removing identity information\footnote{More details are provided in Section 2 of the supplementary material.}. Importantly, since $\mathcal{L}_m^d$ is entropy-based and modality-agnostic, it is applied independently to each modality (visual, physiological, and audio), ensuring that subject-specific identity cues do not interfere with expression-related representations and improving robustness across modalities.

\subsection{Domain Adaptation via Co-training}

\noindent\textbf{(a) Selection of Source Subjects.}\label{sec:method_sss}
Selecting the most relevant source subjects from $\mathcal{S}_m$ is a critical step in \ours method, where the model is trained on diverse sources to ensure robustness. However, when adapting to a target subject $\mathbf{T}_m$ captured in a specific environment and cultural context (e.g., an adult Caucasian male), 
the source subjects must align with these characteristics to ensure effective adaptation. To address this, we leverage our multimodal system to define selection criteria based on \textbf{co-training}, which ensures that the chosen sources are highly relevant to the target subject. 
Given two modalities $visual$ and $physiological$, estimate the cosine similarity between $\mathcal{S}$ and $\mathbf{T}$ using embeddings $\mathbf{h}_m^{\textbf{s}}$ and $\mathbf{h}_m^{\textbf{t}}$,
\begin{equation}
    \mathbf{Q}_m(\mathcal{S}, \mathbf{T}) =\frac{\mathbf{h}_m^\textbf{s} \cdot \mathbf{h}_m^\textbf{t}}{\|\mathbf{h}_m^\textbf{s}\|_2 \cdot \|\mathbf{h}_m^\textbf{t}\|_2}
\end{equation}
where $||.||$ is the euclidean norm calculated for $\mathbf{h}_m^\textbf{s}$ and $\mathbf{h}_m^\textbf{t}$. $\mathbf{Q}_m(\mathcal{S},  \mathbf{T})$ is estimated for each source and target pairs as,
\begin{equation}
\label{eq:src_sel_cos}
\begin{aligned}
\begin{split}
    P_m =[\mathbf{Q}_m^\text{c}(\mathbf{S}_{m_1}, \mathbf{T}_m),\ldots, \mathbf{Q}_v^\text{c}(\mathbf{S}_{m_D}, \mathbf{T}_m)]
\end{split}
\end{aligned}
\end{equation}
Here, $P_m$ represents a dictionary list containing all pairwise distances computed using cosine similarity for each modality. Due to the distinct architectures of the modalities, such as a {2D CNN} for the visual modality and a {1D CNN} for the physiological modality, we normalize $P_m$ to the range $[0, 1]$ to ensure comparability across modalities. We define $P_\text{v}$ and $P_\text{p}$ as the similarity scores for the visual and physiological modalities, respectively. These scores are merged into a combined set $\widehat{P}=(P_\text{v} \bigcup_{}^{}P_\text{p})$, from which we select source subjects whose distances exceed a predefined threshold $\tau_{ss}$. The selection process is formalized as:  
\begin{equation}
\label{eq:close_src}
\begin{aligned}
    \widetilde{\mathcal{S}}_m=\{\mathcal{S}_{m_i}: \widehat{P}_i > \tau_{ss} \}  \quad \forall i \in \left\{ 1,\dots ,D\right\}
\end{aligned}
\end{equation}
The selected source subjects for each modality form a subset $\widetilde{\mathcal{S}}_m$, where $\widetilde{\mathcal{S}}_m \subseteq \mathcal{S}_m$. This selection process ensures that only the most relevant source subjects are used for the target adaptation.

\noindent\textbf{(b) Generation of Target Pseudo-Labels (PLs).} In MSDA, reliable PLs determine the performance of a model on an unseen target subject \cite{zeeshan2024subject}. \ours benefits from the multimodal by selecting confident target PLs using co-training. Given two backbones, $B_\text{v}$ and $B_\text{p}$ 
We introduce samples $\mathbf{x}_{m}^\textbf{t}$ from $\mathbf{T}_m$ to produce softmax probabilities with respect to each modality, $p_\text{v}=\sigma(\mathbf{h}_\text{v}^{\textbf{s}})$ and $p_\text{p}=\sigma(\mathbf{h}_\text{p}^{\textbf{s}})$, taking the maximum probability $\widehat{p}=max(p_\text{v},p_\text{p})$ and applying a predetermined threshold $\tau_{pl}$ to select the confident samples, defined as: 
\begin{equation}
    {\textbf{T}}_m^{l} = \left\{ ({\hat{x}_{m_j},\widehat{y}_{m_j}}) | {\widehat{p} > \tau_{pl}},{j\in {\{1\le j\le N^\textbf{t}\}}}   \right\}
\end{equation} 
\noindent where $\textbf{T}^l_m$ represents the set of confident target samples for $m$ modalities, we also store the confident classes $C={\{{y_i |(x_i,y_i)\in \textbf{T}}_m^{l}\}}$. To differentiate between confident and non-confident samples, we create ${\textbf{T}}^u_m$, defined as: 
\begin{equation}
    {\textbf{T}}_m^{u} = \left\{ ({{x}_{m_j}}) | {\widehat{p}\le \tau_{pl}},{j\in {\{1\le j\le N^\textbf{t}\}}}   \right\}
\end{equation} 
\noindent where $\textbf{T}_m^{u}$, defined the \emph{non-confident} target samples, for the estimation of class-agnostic alignment (Section \ref{sec:method_class_ag}).

\noindent\textbf{(c) Class-Aware Domain Alignment.} We exploit class information to perform intra-class and inter-class alignment across domains. For intra-class alignment, samples from the same class are encouraged to reside in the same distribution space. Conversely, for inter-class alignment, samples from different classes are pushed further apart in the feature space. Class-aware alignment is computed using maximum mean discrepancy (MMD) \cite{sejdinovic2013equivalence}, a statistical measure for estimating the disparity between two distributions. 
To sample out examples from the same class, we perform class-aware sampling on source and target subjects. The classes that belong to $\mathbf{T}^l_m$ were denoted as $\widetilde{C}$. We sample out examples from every subject belonging to multiple labeled sources. In each mini-batch, examples from every source and target are extracted for the same class. The intra-class discrepancy is defined as:
{\small
\begin{equation}
\begin{split}
    D_\text{ita}(\widetilde{\mathcal{S}}_m,\mathbf{T}^l_m)=
    \frac{1}{D}\frac{1}{\widetilde{C}}\sum_{a=1}^{D}\sum_{{c}=1}^{\widetilde{C}}\text{MMD}(\mathbf{h_{m_a}^c},\mathbf{h_m^c})
\end{split}
\end{equation}
}
\noindent where, $\mathbf{h_{m_a}^c}$ extracts features from each source subject, $\mathbf{h_m^c}$ are the features extracted from the target subject, and $\widetilde{C}$ are the confident expression classes. Inter-class is defined as:  
{\small
\begin{equation}
\begin{split}
    D_\text{itr}(\widetilde{\mathcal{S}}_m,\mathbf{T}^l_m)=
    \frac{1}{D}\frac{1}{\widetilde{C}(\widetilde{C}-1)}\sum_{a=1}^{D}\sum_{c=1}^{\widetilde{C}}\sum_{\binom{\acute{c}=1}{\acute{c}\neq c}}^{\widetilde{C}}\text{MMD}(\mathbf{h_{m_a}^c},\mathbf{h}_m^{\acute{c}})
\end{split}
\end{equation} }
\noindent where MMD is computed from the opposite classes $\acute{c}$ for every source with the target. Alignment loss is estimated for every modality (\emph{visual} and \emph{physio}), ensuring that source and target align in their respective distribution space.

\noindent\textbf{(d) Class-Agnostic Domain Alignment.} \label{sec:method_class_ag} Most MSDA methods \cite{kang2020contrastive, scalbert2021multi} employ class-aware alignment to bridge dissimilar distributions, often relying on PLs for the target subject. However, these pseudo-labels are prone to noise, and while denoising techniques have been proposed \cite{scalbert2021multi, xu2018deep, deng2022robust, ren2022multi}, they typically use pre-determined thresholds to retain only the most confident pseudo-labels (CPL). This reliance on thresholds can limit performance, especially in ER, where subtle expressions may lead to discarding useful but less confident samples. To address this, a domain-agnostic loss is introduced that aligns non-confident samples to ensure that all target samples contribute to the adaptation process. This approach maximizes the utilization of available data for improving the target adaptation. The domain-agnostic loss is defined as:
\begin{equation}
    D_\text{agn}(\widetilde{\mathcal{S}}_m,\mathbf{T}^u_m)=\frac{1}{D}\sum_{a=1}^{D}\text{MMD}(\mathbf{h}_{m_a}^\textbf{s},\mathbf{h}_m^{\textbf{t}})
\end{equation}
\noindent where $\mathbf{h}_{m_a}^s$ is the feature vector extracted from $\widetilde{\mathcal{S}}_m$ subjects that contain samples not belonging to $\widetilde{C}$. The total alignment loss for both modalities is calculated as:
{\small
\begin{equation}
\begin{aligned}
\begin{split}
    \mathcal{L}_\text{agn}=
    D_\text{agn}(\widetilde{\mathcal{S}}_\text{v},\mathbf{T}^u_\text{v}) +  D_\text{agn}(\widetilde{\mathcal{S}}_\text{p},\mathbf{T}^u_\text{p})
\end{split}
\end{aligned}
\end{equation}
}
{\small
\begin{equation}
\begin{aligned}
\begin{split}
    \mathcal{L}_\text{aw}=
    [D_\text{ita}(\widetilde{\mathcal{S}}_\text{v},\mathbf{T}^l_\text{v})-D_\text{itr}(\widetilde{\mathcal{S}}_\text{v},\mathbf{T}^l_\text{v})] + \\ 
        [D_\text{ita}(\widetilde{\mathcal{S}}_\text{p},\mathbf{T}^l_\text{p})-D_\text{itr}(\widetilde{\mathcal{S}}_\text{p},\mathbf{T}^l_\text{p})]
\end{split}
\end{aligned}
\end{equation}
}
\noindent where $\mathcal{L}_\text{agn}$ and $\mathcal{L}_\text{aw}$ are class-agnostic and class-aware losses respectively. For each modality, the intra-class and class-agnostic discrepancy is minimized, while the inter-class discrepancy is maximized.

\noindent\textbf{(e) Modality Alignment.}
In multimodal settings, aligning modalities by adding a fusion network has proven to be very successful in enhancing model performance \cite{praveen2022joint, aslam2023privileged}. In \ours method, to align the modalities, a fusion module $F$ is introduced that projects the combined modality embeddings into a shared feature space. To train $F$ for multiple source subjects, we use the embeddings $\mathbf{h}_\text{v}^{\textbf{s}}$ and $\mathbf{h}_\text{p}^{\textbf{s}}$ to perform feature concatenation $\mathbf{h}^{\textbf{s}}=\mathbf{h}_\text{v}^{\textbf{s}} \oplus \mathbf{h}_\text{p}^{\textbf{s}}$, and pass through $F$ to calculate a cross-entropy loss:
\begin{equation}
    \mathcal{L}_\text{}^{\textbf{s}}=\frac{1}{D}\frac{1}{N^{\textbf{s}}}\sum_{i=1}^{D}\sum_{j=1}^{N^{\textbf{s}}}\text{CE}(\mathbf{y}_j^i,F(\mathbf{h}_j^i))
\end{equation}
\noindent where $D$ is the source subject that contains the $N^\textbf{s}$ number of samples.
Modality alignment is estimated for each $\widetilde{\mathcal{S}}$ and $\mathbf{T}$ subject individually. We only calculated for the confident samples for the target $\mathbf{T}^l_m$, the embeddings are extracted  $\mathbf{h}_\text{v}^{	\textbf{t}}$ and $\mathbf{h}_\text{p}^{\textbf{t}}$, and applied concatenation function $\mathbf{h}^{\textbf{t}}=\mathbf{h}_\text{v}^{\textbf{t}} \oplus \mathbf{h}_\text{p}^{\textbf{t}}$, then $F$ is applied to calculate the cross-entropy loss:
\begin{equation}
    \mathcal{L}_\text{unsup}^{\textbf{t}}=\frac{1}{N_l^{\textbf{t}}}\sum_{j=1}^{N_l^{\textbf{t}}}\text{CE}(\widehat{\mathbf{y}}_j,F(\mathbf{h}_j^\textbf{t}))
\end{equation}
\noindent where $N_l^\textbf{t}$ is the total number of reliable samples, and $\widehat{\mathbf{y}}_j$ is the CPL. 
The final objective is to jointly optimize all four losses $\mathcal{L}=\mathcal{L}^\textbf{s} + \gamma\mathcal{L}_\text{unsup}^\textbf{t}+\alpha\mathcal{L}_\text{agn}+\beta\mathcal{L}_\text{aw}$, where $\gamma$, $\alpha$, and $\beta$ hyper-parameters weight the contribution of each loss.

\section{Results and Discussion}
\label{sec:results}

\begin{table*}
\scriptsize
\renewcommand{\arraystretch}{1.4}
\centering
\begin{tabular}{c|c|cccccccccccc}
\hline
\multirow{2}{*}{\textbf{Settings}} & 
\multirow{2}{*}{\textbf{Methods}} & 
\multicolumn{12}{c}{\textbf{Target Subjects}} \\ \cline{3-14}
 & & \textbf{Sub-1} & \textbf{Sub-2} & \textbf{Sub-3} & \textbf{Sub-4} & \textbf{Sub-5} & \textbf{Sub-6} & \textbf{Sub-7} & \textbf{Sub-8} & \textbf{Sub-9} & \textbf{Sub-10} & \textbf{Avg Acc} & \textbf{Avg F1} \\ \hline \hline

Lower Bound &
\begin{tabular}[c]{@{}c@{}}Visual-only \\Physio-only\\Fusion\end{tabular} & 
\begin{tabular}[c]{@{}c@{}}37.1\\41.3\\39.1\end{tabular} & 
\begin{tabular}[c]{@{}c@{}}27.2\\33.2\\35.2\end{tabular} & 
\begin{tabular}[c]{@{}c@{}}26.3\\40.3\\28.6\end{tabular} & 
\begin{tabular}[c]{@{}c@{}}29.6\\34.2\\25.6\end{tabular} & 
\begin{tabular}[c]{@{}c@{}}28.4\\33.4\\39.2\end{tabular} & 
\begin{tabular}[c]{@{}c@{}}28.6\\31.2\\36.6\end{tabular} & 
\begin{tabular}[c]{@{}c@{}}35.6\\37.3\\44.5\end{tabular} & 
\begin{tabular}[c]{@{}c@{}}22.1\\33.9\\27.8\end{tabular} & 
\begin{tabular}[c]{@{}c@{}}33.6\\35.6\\28.8\end{tabular} & 
\begin{tabular}[c]{@{}c@{}}19.7\\25.2\\20.1\end{tabular} & 
\begin{tabular}[c]{@{}c@{}}28.8\\34.6\\32.5\end{tabular} & 
\begin{tabular}[c]{@{}c@{}}20.8\\26.4\\{28.1}\end{tabular} \\ \hline

\begin{tabular}[c]{@{}c@{}}MM-UDA\\[-4pt] (Blending)\end{tabular} &
\begin{tabular}[c]{@{}c@{}}DANN \cite{ganin2016domain} \\
CDAN \cite{long2018conditional} \\
MMD \cite{sejdinovic2013equivalence} \\
\oursuda \end{tabular} & 
\begin{tabular}[c]{@{}c@{}}39.4\\32.6\\34.1\\43.5\end{tabular} & 
\begin{tabular}[c]{@{}c@{}}28.4\\24.1\\26.3\\29.9\end{tabular} & 
\begin{tabular}[c]{@{}c@{}}31.5\\29.2\\31.5\\37.6\end{tabular} & 
\begin{tabular}[c]{@{}c@{}}33.4\\32.1\\36.7\\\textbf{38.2}\end{tabular} & 
\begin{tabular}[c]{@{}c@{}}37.6\\29.6\\34.2\\40.0\end{tabular} & 
\begin{tabular}[c]{@{}c@{}}31.5\\26.8\\33.5\\22.3\end{tabular} & 
\begin{tabular}[c]{@{}c@{}}34.2\\28.4\\43.4\\38.6\end{tabular} & 
\begin{tabular}[c]{@{}c@{}}26.6\\23.6\\34.1\\40.0\end{tabular} & 
\begin{tabular}[c]{@{}c@{}}34.3\\23.4\\28.6\\47.6\end{tabular} & 
\begin{tabular}[c]{@{}c@{}}27.4\\22.6\\31.2\\25.8\end{tabular} & 
\begin{tabular}[c]{@{}c@{}}32.4\\27.2\\33.4\\36.3\end{tabular} & 
\begin{tabular}[c]{@{}c@{}}27.4\\25.3\\28.2\\{28.5}\end{tabular} \\ \hline

MM-MSDA &
\begin{tabular}[c]{@{}c@{}}CAN \cite{kang2020contrastive}\\ 
Sub-based\textsubscript{top-k} \cite{zeeshan2024subject} \\
CMSDA \cite{scalbert2021multi} \\
\oursmsda \end{tabular} &
\begin{tabular}[c]{@{}c@{}}43.2\\43.5\\39.4\\\textbf{49.2}\end{tabular} & 
\begin{tabular}[c]{@{}c@{}}32.2\\42.2\\39.4\\\textbf{43.3}\end{tabular} & 
\begin{tabular}[c]{@{}c@{}}27.2\\28.5\\29.4\\\textbf{47.3}\end{tabular} & 
\begin{tabular}[c]{@{}c@{}}33.6\\22.5\\36.6\\\textbf{38.2}\end{tabular} & 
\begin{tabular}[c]{@{}c@{}}33.4\\29.1\\39.2\\\textbf{40.1}\end{tabular} & 
\begin{tabular}[c]{@{}c@{}}32.5\\32.5\\39.1\\\textbf{44.3}\end{tabular} & 
\begin{tabular}[c]{@{}c@{}}36.6\\40.6\\38.4\\\textbf{46.4}\end{tabular} & 
\begin{tabular}[c]{@{}c@{}}32.4\\41.2\\35.7\\\textbf{45.2}\end{tabular} & 
\begin{tabular}[c]{@{}c@{}}41.5\\38.0\\45.2\\\textbf{50.0}\end{tabular} & 
\begin{tabular}[c]{@{}c@{}}33.8\\28.4\\25.6\\\textbf{34.3}\end{tabular} & 
\begin{tabular}[c]{@{}c@{}}34.6\\34.7\\36.8\\\textbf{43.8}\end{tabular} & 
\begin{tabular}[c]{@{}c@{}}27.3\\28.4\\30.0\\\textbf{35.7}\end{tabular} \\ \hline

Upper Bound & Fine-tuning &
83.6 & 78.5 & 70.0 & 71.6 & 71.5 & 64.5 & 76.8 & 76.9 & 78.5 & 64.8 & 73.6 & 69.1 \\ 
\hline
\end{tabular}
\caption{\centering Accuracy and Average F1 score of \ours with baselines against multimodal UDA and MSDA methods on BioVid. Best Avg Acc and Avg F1 per section are highlighted in bold (ignoring supervised fine-tuning).}
\label{tab:biovid_res}
\vspace{-5pt}
\end{table*}

\begin{table*}
\scriptsize
\renewcommand{\arraystretch}{1.4}
\centering
\begin{tabular}{c|c|cccccccccccc}
\hline
\multirow{2}{*}{\textbf{Settings}} &
\multirow{2}{*}{\textbf{Methods}} &
\multicolumn{12}{c}{\textbf{Target Subjects}} \\ \cline{3-14}
& & \textbf{Sub-1} & \textbf{Sub-2} & \textbf{Sub-3} & \textbf{Sub-4} & \textbf{Sub-5} & \textbf{Sub-6} & \textbf{Sub-7} & \textbf{Sub-8} & \textbf{Sub-9} & \textbf{Sub-10} & \textbf{Avg Acc} & \textbf{Avg F1} \\ \hline \hline

Lower Bound &
\begin{tabular}[c]{@{}c@{}}Visual-only\\Physio-only\\Fusion\end{tabular} &
\begin{tabular}[c]{@{}c@{}}71.1\\57.7\\68.3\end{tabular} &
\begin{tabular}[c]{@{}c@{}}23.2\\81.1\\30.4\end{tabular} &
\begin{tabular}[c]{@{}c@{}}51.4\\60.0\\54.2\end{tabular} &
\begin{tabular}[c]{@{}c@{}}63.3\\28.2\\57.6\end{tabular} &
\begin{tabular}[c]{@{}c@{}}55.6\\60.0\\65.3\end{tabular} &
\begin{tabular}[c]{@{}c@{}}63.2\\69.7\\67.8\end{tabular} &
\begin{tabular}[c]{@{}c@{}}72.7\\84.2\\79.7\end{tabular} &
\begin{tabular}[c]{@{}c@{}}48.7\\51.1\\48.6\end{tabular} &
\begin{tabular}[c]{@{}c@{}}60.0\\34.6\\61.2\end{tabular} &
\begin{tabular}[c]{@{}c@{}}59.6\\29.7\\59.4\end{tabular} &
\begin{tabular}[c]{@{}c@{}}56.9\\55.6\\59.3\end{tabular} &
\begin{tabular}[c]{@{}c@{}}47.9\\34.8\\50.5\end{tabular} \\ \hline

\begin{tabular}[c]{@{}c@{}}MM-UDA\\[-4pt] (Blending)\end{tabular} &
\begin{tabular}[c]{@{}c@{}}DANN \cite{ganin2016domain} \\
CDAN \cite{long2018conditional} \\
MMD \cite{sejdinovic2013equivalence} \\
\oursuda \end{tabular} &
\begin{tabular}[c]{@{}c@{}}65.3\\58.1\\66.3\\\textbf{71.2}\end{tabular} &
\begin{tabular}[c]{@{}c@{}}65.3\\68.1\\52.3\\53.4\end{tabular} &
\begin{tabular}[c]{@{}c@{}}73.3\\72.6\\71.6\\69.7\end{tabular} &
\begin{tabular}[c]{@{}c@{}}66.4\\59.9\\74.5\\70.0\end{tabular} &
\begin{tabular}[c]{@{}c@{}}67.3\\67.9\\71.9\\\textbf{73.5}\end{tabular} &
\begin{tabular}[c]{@{}c@{}}72.3\\70.7\\74.5\\68.7\end{tabular} &
\begin{tabular}[c]{@{}c@{}}73.2\\76.4\\81.2\\82.4\end{tabular} &
\begin{tabular}[c]{@{}c@{}}55.3\\59.5\\57.2\\\textbf{60.0}\end{tabular} &
\begin{tabular}[c]{@{}c@{}}70.4\\60.3\\68.4\\72.7\end{tabular} &
\begin{tabular}[c]{@{}c@{}}57.2\\49.4\\57.2\\61.2\end{tabular} &
\begin{tabular}[c]{@{}c@{}}66.6\\64.2\\67.5\\68.3\end{tabular} &
\begin{tabular}[c]{@{}c@{}}52.9\\52.5\\53.0\\60.7\end{tabular} \\ \hline

MM-MSDA &
\begin{tabular}[c]{@{}c@{}}CAN \cite{kang2020contrastive} \\
Sub-based\textsubscript{top-k} \cite{zeeshan2024subject} \\
\oursmsda \end{tabular} &
\begin{tabular}[c]{@{}c@{}}69.3\\70.8\\\textbf{71.2}\end{tabular} &
\begin{tabular}[c]{@{}c@{}}\textbf{81.2}\\74.9\\\textbf{81.2}\end{tabular} &
\begin{tabular}[c]{@{}c@{}}67.2\\\textbf{73.6}\\\textbf{73.6}\end{tabular} &
\begin{tabular}[c]{@{}c@{}}59.3\\70.6\\\textbf{74.6}\end{tabular} &
\begin{tabular}[c]{@{}c@{}}62.3\\59.8\\\textbf{73.5}\end{tabular} &
\begin{tabular}[c]{@{}c@{}}74.2\\77.9\\\textbf{78.5}\end{tabular} &
\begin{tabular}[c]{@{}c@{}}83.3\\71.8\\\textbf{86.3}\end{tabular} &
\begin{tabular}[c]{@{}c@{}}\textbf{60.0}\\49.9\\\textbf{60.0}\end{tabular} &
\begin{tabular}[c]{@{}c@{}}66.8\\63.9\\\textbf{79.2}\end{tabular} &
\begin{tabular}[c]{@{}c@{}}59.3\\69.8\\\textbf{71.5}\end{tabular} &
\begin{tabular}[c]{@{}c@{}}68.3\\68.3\\\textbf{74.9}\end{tabular} &
\begin{tabular}[c]{@{}c@{}}53.2\\54.1\\\textbf{62.2}\end{tabular} \\ \hline

Upper Bound &
Fine-tuning &
83.4 & 84.5 & 98.2 & 98.1 & 89.7 & 87.6 & 90.0 & 89.6 & 89.4 & 95.7 & {90.6} & {80.4} \\ 
\hline
\end{tabular}
\caption{Accuracy and Average F1 score of \ours with baselines, multimodal UDA and MSDA methods on StressID adapted to 10 unlabeled target subjects. Best Avg Acc and Avg F1 per section are highlighted in bold.}
\label{tab:stressid_res}
\vspace{-5pt}
\end{table*}

\begin{table}[h]
\centering
\scriptsize
\renewcommand{\arraystretch}{1.4}
\setlength{\tabcolsep}{3pt}
\begin{tabular}{c|ccccc|c|c}
\hline
\textbf{Methods} & \textbf{Sub-1} & \textbf{Sub-2} & \textbf{Sub-3} & \textbf{Sub-4} & \textbf{Sub-5} & \textbf{Avg Acc} & \textbf{Avg F1} \\ \hline \hline

Visual-only   & 33.3 & {86.0} & 47.1 & 70.5 & 73.0 & 62.0 & 38.2 \\
Audio-only    & 35.3 & 35.4 & 44.1 & 65.4 & 60.3 & 48.1 & 36.1 \\
Fusion        & 36.8 & 81.7 & 47.6 & {73.6} & {77.6} & 63.5 & {38.4} \\ \hline

MMD~\cite{sejdinovic2013equivalence}           & 29.9 & 81.0 & {53.4} & 68.8 & 77.6 & 62.1 & {38.8} \\
\oursuda    & 35.4 & {86.6} & 46.7 & 73.5 & {79.4} & {64.3} & 38.7 \\ \hline

Sub-based\textsubscript{top-k}~\cite{zeeshan2024subject}     & 33.4 & 84.5 & \textbf{54.2} & 71.4 & 79.5 & 64.6 & 38.9 \\
\oursmsda   & \textbf{39.7} & \textbf{95.0} & \textbf{54.2} & \textbf{79.3} & \textbf{81.2} & \textbf{69.9} & \textbf{41.2} \\ \hline

\end{tabular}
\caption{Performance of \ours against baselines, multimodal UDA, and MSDA methods on the BAH dataset.}
\label{tab:bah_res}
\vspace{-15pt}
\end{table}


\subsection{Experimental Protocol}

\textbf{Datasets.} 
\ours was evaluated on three challenging multimodal ER datasets for subject-based adaptation. \textbf{BioVid Heat Pain (Part A)} \cite{walter2013biovid} contains 87 subjects with five pain levels; following \cite{zeeshan2024subject}, we use 77 as sources and 10 as targets. \textbf{StressID} \cite{chaptoukaev2023stressid} comprises 65 participants in controlled settings. For our experimental setup, 44 source and 10 target subjects were used with both visual and physiological (EDA) modalities. \textbf{Behavioural Ambivalence/Hesitancy (BAH)~\cite{gonzalez2025bah}} consists of 224 participants from diverse demographics in uncontrolled, real-world settings. Our analysis utilizes 143 of these individuals as source subjects and five as targets with visual and audio modalities.

\noindent\textbf{Implementation Detail.}
In addition to the visual modality, the physiological modality, specifically electrodermal activity (EDA), is incorporated for both the BioVid and StressID datasets. For visual, ResNet18 \cite{he2016deep} is used, following the same protocol as \cite{zeeshan2024subject}. Additional experiments using an alternative visual backbone are reported in the suppl material. For the physio modality, an LSTM-based 1D-CNN network, and for the audio modality, PANN with CNN14~\cite{kong2020panns} model is used for the BAH dataset. All the target subjects are split into train/val/test sets, and all the results are reported using a target test set. For selection of source subjects, $\tau_{ss}$ is set to 0.55 (Sec.~\ref{sec:ablation}-b). Following prior work~\cite{deng2022robust, scalbert2021multi, ren2022multi}, we set the initial $\tau_{pl}$ to 0.95 and gradually decrease it by $0.01$ every $N$ epochs to incorporate more samples as training stabilizes, with this choice validated as well as further datasets and implementation details, provided in the suppl material.


\noindent\textbf{Baseline Methods.} 
The reported results correspond to the top-1\% multimodal target accuracy across all target subjects. The evaluation begins with the \textbf{Lower Bound}, where a model is trained solely on labeled source subjects and directly evaluated on the target test set without any target supervision. Results are reported for \emph{visual-only}, \emph{physio-only}, or \emph{audio-only}, and \emph{fusion} (vision+physio) or (vision+audio) modalities. Next, the \textbf{MM-UDA (blending)} setting blends all the source subjects into a single domain, then adapts to the unlabeled target subject using various UDA techniques, including \emph{DANN} \cite{ganin2016domain}, \emph{CDAN} \cite{long2018conditional}, \emph{MMD} \cite{sejdinovic2013equivalence}, and \oursuda, which only use co-training to generate target pseudo-labels. Furthermore, \oursmsda is compared against \textbf{MM-MSDA} approaches, including \emph{CAN} \cite{kang2020contrastive}, \emph{Sub-based MSDA}, and \emph{CMSDA} \cite{scalbert2021multi}. In all methods, we use the same backbones for every modality to ensure a fair comparison with our proposed approach. Finally, results for an \emph{Upper Bound} are reported, where models are fine-tuned using labeled data from the target subject.

\subsection{Comparison with State-of-the-Art Methods
}

Tab.~\ref{tab:biovid_res} shows results on the \textbf{BioVid} dataset. All domain adaptation approaches outperform the lower bound (source-only) baselines (visual and both), confirming the effectiveness of multimodal domain adaptation in the context of pain estimation. Notably, the \emph{physio-only} baseline outperforms many standard UDA and MSDA methods, achieving an average accuracy of 34.5\%. This reflects findings in prior work \cite{werner2017analysis, waligora2024joint, aslam2024distilling}, where physio signals are shown to be strong indicators of pain intensity, especially in subtle expressions where visual cues are often weak or ambiguous. Similarly, our \emph{visual-only} and \emph{fusion} baselines yield weaker accuracy than \emph{physio-only}. Nonetheless, \emph{fusion} improves recall on harder classes, resulting in a higher Avg F1 score (28.1\%) compared to \emph{physio-only} (26.4\%), despite its slightly lower accuracy.
Among all methods, our approach achieves the highest overall performance. \oursuda outperforms both the \emph{lower-bound} and \emph{UDA} baselines and is competitive with MSDA methods. Further, \oursmsda achieves even better results, with a 9.3\% gain over \emph{physio-only} and 7.5\% over \oursuda. It also outperforms MSDA baselines, with an average gain of 9.2\% over \emph{CAN} \cite{kang2020contrastive} and \emph{Sub-based\textsubscript{top-k}} \cite{zeeshan2024subject}, and 7\% over \emph{CMSDA} \cite{scalbert2021multi}. This shows our method effectively exploits modality complementarity, adapting to subject-specific signals in subtle pain recognition.
\begin{figure*}[t!]
\centering
\includegraphics[width=1.0\linewidth]{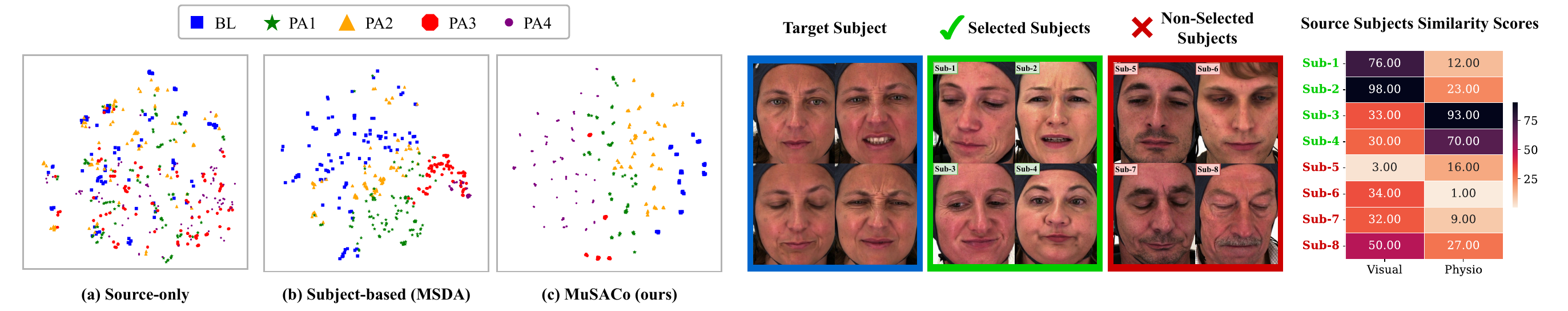}
\caption{\textbf{Left:} t-SNE visualizations. The source-only produced indistinguishable feature class clusters. Sub-based MSDA \cite{zeeshan2024subject} reduces some noise by separating classes to some extent. \ours creates more separable clusters for each class. \textbf{Right:} Visualization of selected and non-selected source subjects for a reference target subject, with similarity scores from visual and physiological modalities.}
\label{fig:visualization}      
\vspace{-10pt}
\end{figure*}
Tab.~\ref{tab:stressid_res} presents the performance on the \textbf{StressID} dataset. As expected, training without adaptation (lower bound) yields significantly lower accuracy across most subjects. Notably, fusing both modalities in the lower bound produces more stable performance compared to using either modality alone. This is due to the complementary nature of the modalities; when the visual modality underperforms, the physio often compensates, resulting in more balanced performance across subjects. Among existing methods, \emph{CAN} achieves the highest performance on {Sub-2, 8}, whereas \emph{Sub-based\textsubscript{top-k}} gains the performance on {Sub-3} that is comparable with our method. \oursuda improved the performance on 3 out of 10 subjects while, on average, attaining a similar performance as MSDA methods. \oursmsda outperformed every other method on all target subjects, with an overall gain of 15.7\% from \emph{lower-bound (both)}, 7.4\% from \emph{MMD}, 6.7\% from \oursuda and \emph{CAN}, and 6.6\% from \emph{Sub-based MSDA}. Tab.~\ref{tab:bah_res} reports results on the \textbf{BAH} dataset. It can be seen that the visual modality is much stronger than audio, but fusion leverages complementary information across both modalities to achieve slight gains in performance over unimodal baselines. Among the UDA methods, \emph{MMD} improves over the lower bound, while \oursuda further stabilizes adaptation across subjects. In the MSDA setting, \emph{Sub-based} demonstrates competitive results; however, \oursmsda consistently outperforms all baselines, surpassing the lower-bound fusion by 6.4\% in average accuracy and 2.3\% in F1 score. Subject-wise F1 scores were provided in suppl. material.

\subsection{Visualizations} 
In Fig.~\ref{fig:visualization} (left), t-SNE \cite{van2008visualizing} shows how different methods separate class embeddings. The source-only model yields noisy, overlapping features. Subject-based MSDA \cite{zeeshan2024subject} improves clustering but struggles with overlaps (e.g., \emph{PA3} in \emph{PA4}). In contrast, \ours better clusters same-class samples and separates opposing ones.
Fig.~\ref{fig:visualization} (right) shows the closest and furthest source subjects to a target (\emph{Sub-10}, BioVid), based on similarity scores from visual and physiological modalities. A source is selected if either modality exceeds the threshold $\tau_{ss}$, e.g., \emph{Sub-1/2} (visual) and \emph{Sub-3/4} (physio). \emph{Sub-5} to \emph{Sub-8} are excluded as both scores fall below $\tau_{ss}$ (see Sec.~\ref{sec:ablation}(b) for threshold selection).

\subsection{Ablation Studies}
\label{sec:ablation}

\begin{figure}[t!]
\centering
\includegraphics[width=1.02\linewidth]{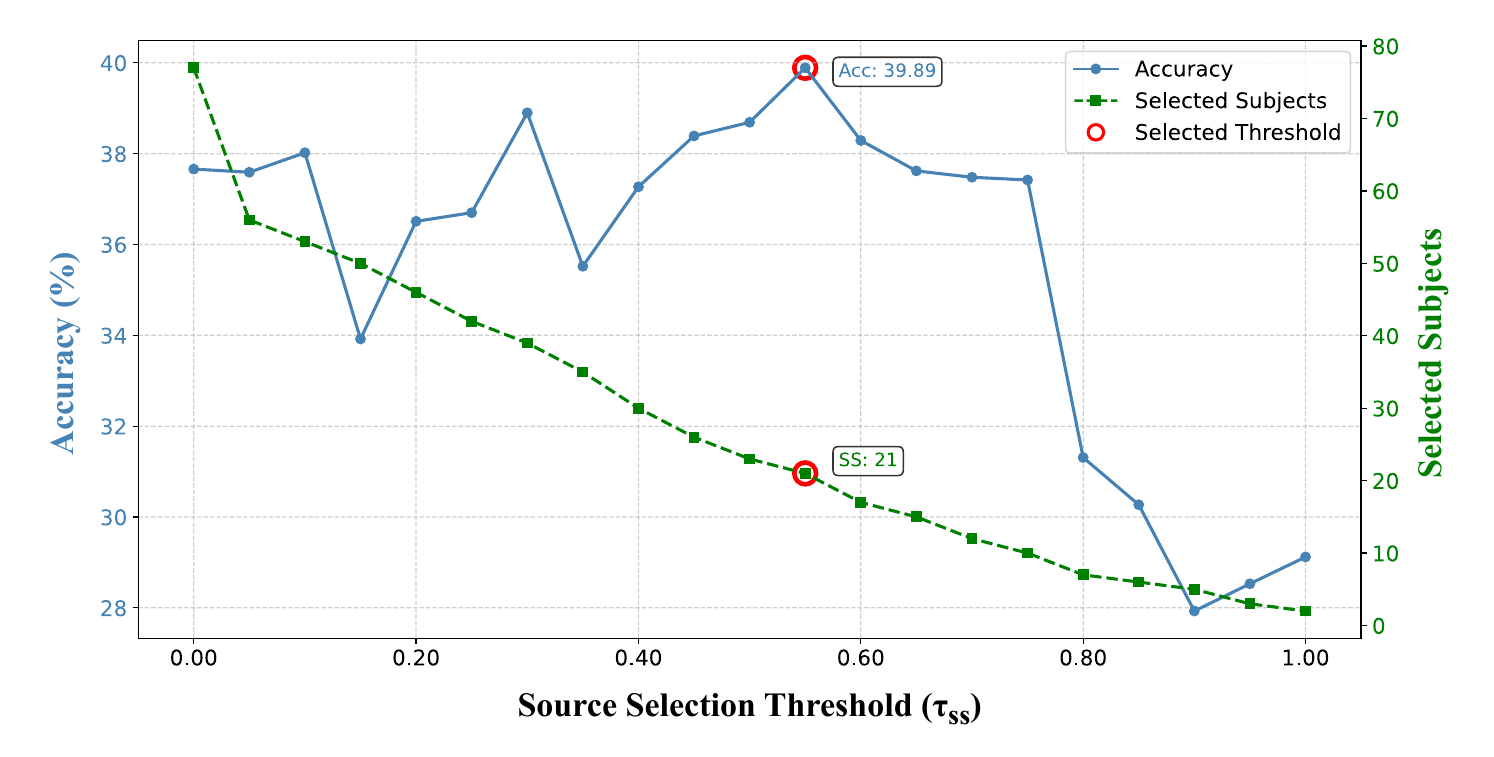}
\caption{Source subject threshold ($\tau_{ss}$) selection based on the accuracy. The \textcolor{red}{red circle} highlights the selected $\tau_{ss}=0.55$.}
\label{fig:ss_threshold}      
\vspace{-15pt}
\end{figure}

\noindent\textbf{(a) Impact of difference loss components.}
An incremental ablation study is conducted by gradually introducing loss terms under the same configuration and weighting hyperparameters as the final model. Tab.~\ref{tab:loss_and_tau} shows that adding the target loss $\mathcal{L}^t_{unsup}$ significantly improves Accuracy and F1 scores over the source-only baseline. Further gains are achieved by incorporating the class-aware loss $\mathcal{L}_{aw}$ and class-agnostic loss $\mathcal{L}_{agn}$, both of which contribute to improving the adaptation process in the absence of target labels. The combination of all loss terms achieves the highest performance, reaching 43.8\% Accuracy and 35.7\% F1.
\noindent\textbf{(b) Impact of selection of source subjects threshold.} 
In Fig.~\ref{fig:ss_threshold}, setting $\tau_{ss}$=0 disables the selection mechanism, including all source subjects (77 for Biovid), achieving an average accuracy of 37.6\%. As $\tau_{ss}$ increases, fewer but more relevant source subjects are selected based on similarity scores, improving performance. The best accuracy of 39.8\% is achieved at $\tau_{ss}$=0.55, where only 21 source subjects are selected on average. This shows the effectiveness of co-training in identifying complementary informative sources for better target adaptation. Subject-wise results are provided in the suppl. material.



\begin{table}[h]
\centering
\footnotesize
\renewcommand{\arraystretch}{1.3}
\begin{tabular}{cccc|c|c}
\hline
$\mathcal{L}^s$ & $\mathcal{L}^t_{unsup}$ & $\mathcal{L}_{aw}$ & $\mathcal{L}_{agn}$ & \textbf{Avg Acc} & \textbf{Avg F1} \\
\hline \hline
\checkmark &  &  &  & 32.5 & 28.1 \\
\checkmark & \checkmark &  &  & 36.6 & 29.2 \\
\checkmark & \checkmark & \checkmark & & 38.1 & 29.7 \\ 
\checkmark & \checkmark &  & \checkmark  & 39.3 & 31.2 \\
\checkmark & \checkmark & \checkmark & \checkmark & \textbf{43.8} & \textbf{35.7} \\
\hline
\end{tabular}
\caption{Impact of individual loss components. Average performance across all subjects of the BioVid dataset.}
\label{tab:loss_and_tau}
\vspace{-2em}
\end{table}

\section{Conclusion}
\label{sec:conclusion}
Domain adaptation methods typically focus on individual modalities, or in the case of multimodal MSDA, they aggregate data from all source subjects, losing critical subject-specific cues necessary for effective adaptation to unlabeled target data. In this paper, \ours, a multimodal subject-specific MSDA is introduced that effectively leverages subject-level information for a personalized adaptation. In particular, \ours selects the most similar source subjects based on multimodal similarity and applies co-training to generate PLs for the target. Class-aware and class-agnostic losses are combined to leverage both confident and uncertain target samples effectively. Final alignment is achieved via feature concatenation across modalities. \ours is validated on challenging MER datasets, showing consistent gains across methods. \ours supports personalized modeling by adapting to each target subject through relevant sources, making it well-suited for health-related applications. Future work will explore test-time adaptation to reduce per-subject retraining and enable faster subject-specific adaptation. 
\section*{Acknowledgment} This research was partially supported by the Natural Sciences and Engineering Research Council of Canada, Fonds de recherche du Québec – Santé, Canada Foundation for Innovation, and the Digital Research Alliance of Canada.

\section{Supplementary}
\label{sec:Supplementary}

\section{Additional Implementation Details}

{For visual, ResNet18 \cite{he2016deep} network, physio network based on an LSTM-based 1D-CNN network,
consisting of two convolutional layers, one LSTM layer, and one fully connected layer. For the audio modality, we employ the Pretrained Audio Neural Network (PANNs)~\cite{kong2020panns} framework, specifically the CNN14 architecture, pretrained on AudioSet. Additionally, the expression head is constructed using two 2-MLP layers for each network. A batch size of 32 for the target subject with images resized to 100$\times$100 resolution, and the model is trained for 20 epochs.}

\subsection{Algorithm}
\noindent\textbf{(a) Selection of Subject-Specific Sources.}
Algorithm \ref{algo:sub_sel} shows the subject-specific selection of source subjects with co-training. Given source subjects $\mathcal{S}$ and specific target subject $\mathbf{T}$, extracted features using visual $B_\text{v}$ and physiological $B_\text{p}$ encoders. Construct a similarity metric $z^v$ and $z^p$ by measuring CosineSimilarity between every source subject and target. To estimate the maximum score, we normalize the similarities, merge them, and sort them in descending order, followed by a threshold that selects the most relevant subjects from the target.       

\begin{algorithm}
\caption{Subject-Specific Selection of Source Subjects}
\label{algo:sub_sel}
\begin{algorithmic}[0]
\Require 
\State $\mathcal{{S}}$: labeled source subjects, $\mathbf{T}$: unlabeled target domain, $B_\text{vis}$: visual encoder, $B_\text{phy}$: physio encoder, $\tau_{ss}$: threshold to select relevant source subjects
\end{algorithmic}
\begin{algorithmic}[1]
\State \textbf{Initialize:} $\mathbf{z}^v \gets []$, $\mathbf{z}^p \gets []$
\State { Decompose $\mathbf{T}$ into $\mathrm{T}^v$ and $\mathrm{T}^p$}
\State {\color{blue}{\# Extract target features}}
\State $\mathrm{X}_t^v \gets {B}_{vis}(\mathrm{T}^v)$, $\mathrm{X}_t^p \gets {B}_{phy}(\mathrm{T}^p)$
\For{each domain $\mathbf{S}_i \in \mathcal{S}$}
    \State \parbox[t]{\dimexpr\linewidth-\algorithmicindent}{Decompose $\mathbf{S}_i$ into $\mathrm{S}^v$ and $\mathrm{S}^p$}
    \State {\color{blue}{\# Extract source features}}
    \State $\mathrm{X}_s^v \gets {B}_{vis}(\mathrm{S}^v)$,
    $\mathrm{X}_s^p \gets {B}_{phy}(\mathrm{S}^p)$ 
    
    \State {\color{blue}{\# Compute similarities}}
    \State $\mathrm{z}_i^v \gets \cos(\mathrm{X}_s^v, \mathrm{X}_t^v)$ 
    \State $\mathrm{z}_i^p \gets \cos(\mathrm{X}_s^p, \mathrm{X}_t^p)$
    \State {\color{blue}{\# Append similarities}}
    \State $\mathbf{z}^v \gets \mathbf{z}^v \cup \mathrm{z}_i^v$, $\mathbf{z}^p \gets \mathbf{z}^p \cup \mathrm{z}_i^p$
\EndFor
\State $\mathbf{z}^v \gets \mathrm{norm}(\mathbf{z}^v)$, $\mathbf{z}^p \gets \mathrm{norm}(\mathbf{z}^p)$
\State Merge similarities: $\mathbf{z} \gets \mathrm{merge}(\mathbf{z}^v, \mathbf{z}^p)$
\State Sort $\mathbf{z}$ in descending order
\State Select relevant sources : $ \widetilde{\mathcal{S}} \gets \tau_{ss}(\mathbf{z})$
\end{algorithmic}
\label{alg:source_selection}
\end{algorithm}

\noindent\textbf{(b) Training Protocol of \ours.}
In algorithm \ref{algo:train_msaco}, we show our training protocol of adapting to an unlabeled target subject. All the equation numbers referenced here correspond to those in the main paper.

\begin{algorithm}
\caption{Training protocol of MSACo}
\label{algo:train_msaco}
\begin{algorithmic}[0]
\Require 
\State $\widetilde{\mathcal{{S}}}$: selected labeled source subjects
\State $\mathbf{T}$: unlabeled target domain
\end{algorithmic}
\begin{algorithmic}[1]

\For{epoch}

\State \parbox[t]{\dimexpr\linewidth-\algorithmicindent}{
Perform co-training on $\mathbf{T}$ to generate pseudo-labels (after every \textit{n} epochs) 
}

\For{iteration}
\State {\color{blue}{\# Class-aware alignment}}
\State \parbox[t]{\dimexpr\linewidth-\algorithmicindent}{ Class-aware domain sampling of $\widetilde{\mathcal{S}}$ and $\mathbf{T}$
}
\State Estimate intra-class discrepancy using (Eq: 8)
\State Estimate inter-class discrepancy using (Eq: 9)
\State Compute class-aware loss $\mathcal{L}_{aw}$ (Eq: 12)
\State {\color{blue}{\# Domain-agnostic alignment}}
\State Domain-agnostic sampling of $\widetilde{\mathcal{S}}$ and $\mathbf{T}$
\State Estimate domain agnostic using (Eq 10)
\State Compute domain-agnostic loss $\mathcal{L}_{agn}$ (Eq: 11)

\State {\color{blue}{\# Modality alignment}}
\State \parbox[t]{\dimexpr\linewidth-\algorithmicindent}{ Perform feat concatenation of source subjects between modalities $\mathbf{h}_\text{v}^{\textbf{s}}$ and $\mathbf{h}_\text{p}^{\textbf{s}}$ \\
}
\State \parbox[t]{\dimexpr\linewidth-\algorithmicindent}{ Perform feat concatenation of target subject between modalities $\mathbf{h}_\text{v}^{\textbf{t}}$ and $\mathbf{h}_\text{p}^{\textbf{t}}$ \\
}

\State Compute modality align loss $\mathcal{L}_\text{}^{\textbf{s}}$ using (Eq: 13)
\State Compute modality align $\mathcal{L}_\text{unsup}^{\textbf{t}}$ using (Eq: 14)
\EndFor

\EndFor

\end{algorithmic}
\end{algorithm}





\subsection{Hyperparameters}
\textbf{Training of backbones.} momentum=0.9, weight decay=5e-4, stochastic gradient descent (SGD) optimizer \cite{ruder2016overview}, learning rate=1e-4 with lr scheduler (eta\_min=0.00002). 
\textbf{Weighting parameters.} In \ours, we give different weights to different loss functions. For the contribution of class-agnostic loss $\gamma=0.5$, and for class-aware loss $\alpha=0.1$.
\textbf{Disentanglement.} Knife is very sensitive to hyperparameters; we have explored several parameters to make it work with the expression recognition task. The most critical parameters that is selected for our experiments are:  zd\_dim=1024, zc\_dim=77, hidden\_state=512, layers=3, nb\_mixture=10, with learning-rate=0.01.

\section{Detail on Training the Source Backbones}
\label{sec:sup_detail_disen}
\begin{figure}[t!]
\centering
\includegraphics[width=0.99\linewidth]{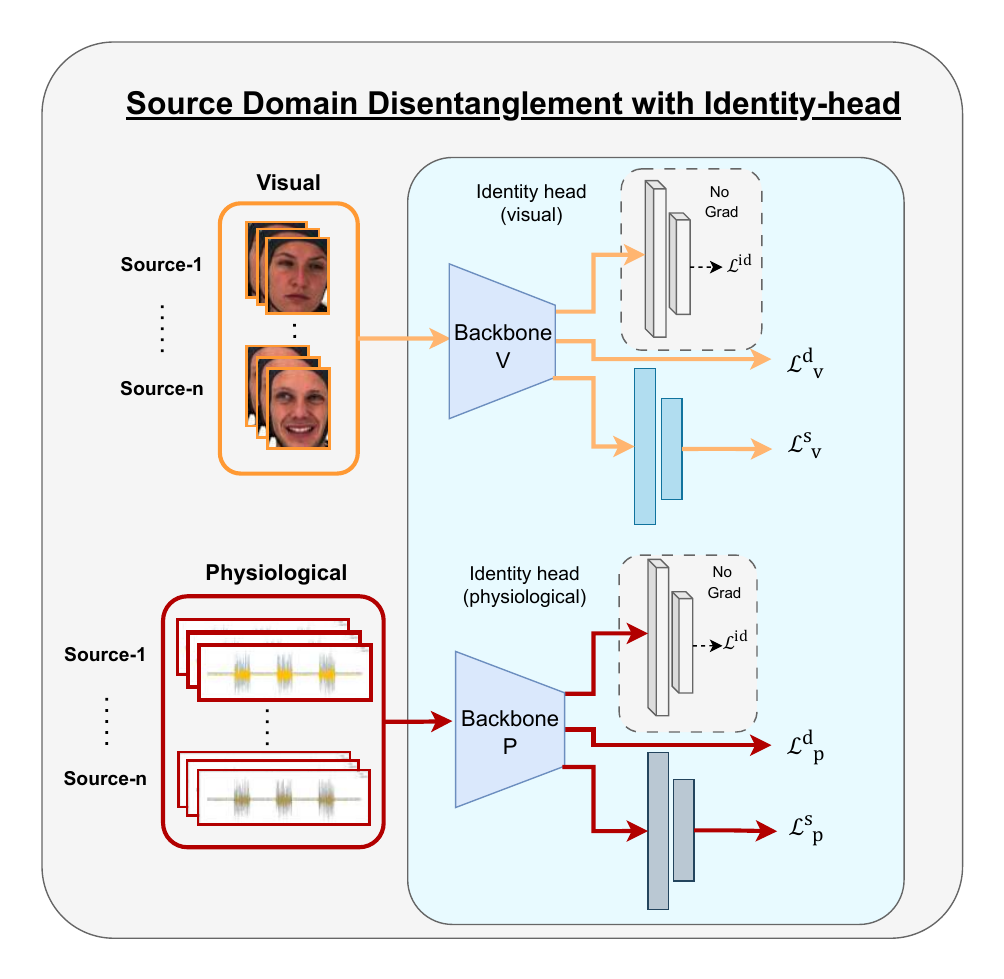}
\caption{Training of disentanglement with knife-loss and identity-head.}
\label{fig:dimsa_network}      
\vspace{-10pt}
\end{figure}
Training of the source backbones (visual and physiological) involves disentangling the identity-related features from the expression task. Our method is inspired by KNIFE\cite{pichler2022differential}, a fully differentiable entropy estimator, which we adapted for disentanglement in a multi-modal framework. The KNIFE estimator optimizes the backbones (visual and physiological) by decoupling non-task-related information through individual modality-specific gradient-based optimization. We leverage the KNIFE estimator to minimize the mutual information between modalities in eliminating identity features and enhancing the disentanglement of task-relevant features across distributions. We first estimate the marginal entropy of embedding $\mathbf{h}_m^\textbf{s}$ as,
\begin{equation}
H(\mathbf{h}_m^\textbf{s})=-\mathbb{E}[MM{p}(\mathbf{h}_m^{\textbf{s}})]
\end{equation}
where $\mathbb{E}$ is the expectation over the distribution $\mathbf{h}_m^\textbf{s}$ and $p(.)$ is the probability density. 
To estimate the conditional entropy between the features and identities, For each sample in the source subjects, we convert the identity labels $\acute{y}_{m_j}^\textbf{s}$ into one-hot representations $\acute{Y}_{m_j}^\textbf{s} \in R^{k}$. Then, conditional entropy is calculated as,
\begin{equation}
H(\mathbf{h}_{m_j}^\textbf{s}|\acute{Y}_{m_j}^{s})=-\mathbb{E}[\log{p}(\mathbf{h}_{m_j}^\textbf{s}|\acute{Y}_{m_j}^\textbf{s})]
\end{equation}
where $p(.,.)$ is the conditional probability density between features $\mathbf{h}_{m_j}^\textbf{s}$ and prediction $\acute{Y}_{m_j}^\textbf{s}$. The total learning loss is calculated as: 
\begin{equation}
\mathcal{L}_{m}^d=H(\mathbf{h}_m^\textbf{s})+H(\mathbf{h}_{m_j}^\textbf{s}|\acute{Y}_{m_j}^\textbf{s}).
\end{equation}  
By minimizing $\mathcal{L}_{m}^d$, the model is encouraged to decouple the information associated with $\acute{y}_m^\textbf{s}$ from the feature embeddings $h_m^\textbf{s}$. Furthermore, we introduce a fixed Identity-head $I_m$ head composed of two fully connected layers, as shown in Figure \ref{fig:dimsa_network}. It takes embeddings $\mathbf{h}_m^\textbf{s}$ without gradient backpropagation. $I_m$ works as a regularizer in conjunction with the disentanglement loss to suppress non-discriminative identity-related information during target adaptation. The disentanglement loss decouples identity-related information, while the $I_m$ helps to constrain redundant (non-task-specific) features, ensuring the model focuses on learning relevant expression-related representations.

\section{Datasets}
\subsection{Biovid Heat Pain} \cite{walter2013biovid} dataset consists of 87 subjects captured in a controlled environment. It consists of five classes, including: \emph{No-Pain}, \emph{PA-1}, \emph{PA-2}, \emph{PA-3}, and \emph{PA-4}. Every individual recorded 20 videos per class, corresponding to 100 videos per subject. We follow the same protocol as \cite{werner2017analysis}, where it eliminates the initial 2 seconds from the video, which does not show any spike indicating pain. In our experiment, we follow \cite{zeeshan2024subject} to consider 77 subjects as sources and 10 subjects as targets: Sub-1 (081014\_w\_27), Sub-2 (101609\_m\_36), Sub-3 (112009\_w\_43), Sub-4 (091809\_w\_43), Sub-5 (071309\_w\_21), Sub-6 (073114\_m\_25), Sub-7 (080314\_w\_25), Sub-8 (073109\_w\_28), Sub-9 (100909\_w\_65), Sub-10 (081609\_w\_40). The label distribution statistics for the test set of Biovid were shown in Tab.~\ref{tab:biovid_dist}. Each subject has nearly balanced distributions across the five classes (neutral + 4 pain levels), confirming the dataset is overall class-balanced.

\subsection{StressID} \cite{chaptoukaev2023stressid} dataset comprised 65 (47 men and 18 women) subjects taken in a lab-controlled environment. Each participant was exposed to 11 tasks, grouped into 4 categories: watching emotional videos, breathing, interactive tasks, and relaxation. These tasks were captured using three different modalities (visual, physiological, and audio). Due to the unavailability of certain modalities in 11 of the participants, we only consider 54 individuals, which includes all the modalities. In our experiments, we take 44 subjects as sources and 10 subjects as targets. The target subject selection follows a 70\% (men) and 30\% (women) ratio, reflecting the higher number of male participants in the dataset. For each participant, there are up to 11 task-specific videos, from which we extracted frames at 1 fps for per-image expression classification. Tab.~\ref{tab:stressid_demo} shows the selected target subject demographics (ID, Gender) and the total number of samples. Tab.~\ref{tab:stressid_dist} shows the label distribution of StressID datasets. It exhibits a noticeable imbalance between stress and no-stress samples across subjects. Only two subjects, Sub-1 and Sub-8, are somewhat balanced, while others are skewed.

\subsection{Behavioural Ambivalence/Hesitancy (BAH)}
The Behavioural Ambivalence/Hesitancy (BAH)~\cite{gonzalez2025bah} dataset is the first multimodal resource designed for subject-specific recognition of ambivalence and hesitancy (A/H) in real-world settings. It consists of 1,118 videos (8.26 hours, including 1.5 hours of A/H) from 224 participants across 9 Canadian provinces, covering diverse ages and ethnicities. Participants answered 7 questions, some explicitly eliciting A/H, while being recorded via webcam and microphone, providing multimodal signals such as facial expressions, vocal cues, and transcripts. Expert annotations mark A/H occurrences at both frame- and video-level, with aligned face crops, transcripts, and participant metadata also included. Unlike controlled lab datasets (e.g., BioVid, StressID), BAH captures natural, in-the-wild behaviours, making it uniquely challenging and valuable for multimodal domain adaptation. In our experiments, we use 143 training data from BAH as subjects to consider as source domains and use 5 subjects from the test-set of BAH as targets: Sub-1 (82711), Sub-2 (82585), Sub-3 (82683), Sub-4 (82708), and Sub-5 (82632), enabling a realistic evaluation of \ours under noisy, heterogeneous conditions. The label distribution statistics for the test set of Stress-ID were shown in Tab.~\ref{tab:bah_dist} It also shows a significant imbalance in ambivalence vs. non-ambivalence across most subjects, with only Sub-1 (82711) and Sub-8 (82683) being closer to balanced.

\begin{table}[h]
\centering
\scriptsize
\renewcommand{\arraystretch}{1.3}
\begin{tabular}{c|c|ccccc|c}
\hline
\textbf{Subjects} & \textbf{ID} & \textbf{NA} & \textbf{PA1} & \textbf{PA2} & \textbf{PA3} & \textbf{PA4} & \textbf{Total} \\
\hline \hline
Sub-1  & 081014\_w\_27 & 287 & 285 & 311 & 305 & 312 & 1500 \\
Sub-2  & 101609\_m\_36 & 325 & 278 & 323 & 274 & 300 & 1500 \\
Sub-3  & 112009\_w\_43 & 304 & 282 & 296 & 310 & 308 & 1500 \\
Sub-4  & 091809\_w\_43 & 265 & 316 & 295 & 298 & 302 & 1476 \\
Sub-5  & 071309\_w\_21 & 299 & 297 & 295 & 304 & 297 & 1492 \\
Sub-6  & 073114\_m\_25 & 295 & 310 & 320 & 283 & 292 & 1500 \\
Sub-7  & 080314\_w\_25 & 311 & 318 & 288 & 276 & 307 & 1500 \\
Sub-8  & 073109\_w\_28 & 314 & 300 & 300 & 272 & 303 & 1489 \\
Sub-9  & 100909\_w\_65 & 298 & 298 & 313 & 279 & 312 & 1500 \\
Sub-10 & 081609\_w\_40 & 300 & 297 & 313 & 286 & 303 & 1499 \\
\hline
\end{tabular}
\caption{Label distribution for BioVid dataset across target subjects test-set.}
\label{tab:biovid_dist}
\end{table}


\begin{table}
  \centering
 \small
  \renewcommand{\arraystretch}{1.3}
  \begin{tabular}{c|c|lccc}
    \hline
    \textbf{Subjects} & \textbf{ID} & \textbf{Gender} & \textbf{NSV} & \textbf{SV} & \textbf{No. frames} \\
    \hline\hline
    Sub-1 & kycf & Men & 5 & 6 & 1021 \\
    Sub-2 & uymz & Men & 8 & 3 & 880 \\
    Sub-3 & h8s1 & Men & 4 & 7 & 957 \\
    Sub-4 & ctzy & Men & 2 & 9 & 1041 \\
    Sub-5 & p9i3 & Men & 5 & 6 & 1041 \\
    Sub-6 & 7h5u & Men & 5 & 5 & 917 \\
    Sub-7 & g7r2 & Men & 8 & 3 & 1041 \\
    Sub-8 & b9w0 & Women & 4 & 7 & 1026 \\
    Sub-9 & r3zm & Women & 3 & 8 & 909 \\
    Sub-10 & x1q3 & Women & 2 & 9 & 1001 \\
    \hline
  \end{tabular}
    \caption{StressID target subject demographics and total number of samples (NSV: No Stress Videos, SV: Stress Videos)}
     \label{tab:stressid_demo}
\end{table}

\begin{table}[h]
\centering
\small
\renewcommand{\arraystretch}{1.3}
\begin{tabular}{c|c|ccc}
\hline
\textbf{Subjects} & \textbf{ID} & \textbf{No-Stress} & \textbf{Stress} & \textbf{Total} \\
\hline\hline
Sub-1  & kycf & 118 & 86  & 204 \\
Sub-2  & uymz & 148 & 28  & 176 \\
Sub-3  & h8s1 & 128 & 83  & 211 \\
Sub-4  & ctzy & 60  & 148 & 208 \\
Sub-5  & p9i3 & 125 & 83  & 208 \\
Sub-6  & 7h5u & 127 & 57  & 184 \\
Sub-7  & g7r2 & 175 & 33  & 208 \\
Sub-8  & b9w0 & 105 & 100 & 205 \\
Sub-9  & r3zm & 63  & 120 & 183 \\
Sub-10 & x1q3 & 59  & 142 & 201 \\
\bottomrule
\end{tabular}
\caption{Label distribution for StressID dataset across target subjects test-set.}
\label{tab:stressid_dist}
\end{table}

\begin{table}[h]
\centering
\small
\renewcommand{\arraystretch}{1.3}
\begin{tabular}{c|c|ccc}
\hline
\textbf{Subjects} & \textbf{ID} & \textbf{No-Amb} & \textbf{Amb} & \textbf{Total} \\
\hline\hline
Sub-1  & 82711 & 433  & 874 & 1307 \\
Sub-2  & 82687 & 319  & 207 & 526  \\
Sub-3  & 82585 & 575  & 85  & 660  \\
Sub-4  & 82592 & 737  & 93  & 830  \\
Sub-5  & 82598 & 1199 & 189 & 1388 \\
Sub-6  & 82632 & 854  & 243 & 1097 \\
Sub-7  & 82681 & 354  & 207 & 561  \\
Sub-8  & 82683 & 445  & 456 & 901  \\
Sub-9  & 82708 & 249  & 104 & 353  \\
Sub-10 & 82714 & 231  & 234 & 465  \\
\bottomrule
\end{tabular}
\caption{Label distribution for BAH dataset across target subjects test-set.}
\label{tab:bah_dist}
\end{table}

\section{Ablation}

\subsection{Impact of Weighting Hyperparameters.}
We performed a sequential weight sensitivity analysis for the key loss terms, $\mathcal{L}_{unsup}^t$, $\mathcal{L}_{aw}$, and $\mathcal{L}_{agn}$ shown in Tab.~\ref{tab:loss_weight_ablation}). The weight sensitivity analysis was performed on three representative subjects from the BioVid dataset: a young woman (081014\_w\_27), a young man (073114\_m\_25), and an older woman (100909\_w\_65). The reported results were the average of all three subjects, and each experiment was run for 10 epochs. These subjects were chosen to reflect the diversity of the dataset across age and gender groups. First, we varied the weight of the target PL loss while keeping the others fixed at 1. Using the best result, we tuned the class-aware alignment loss, followed by the class-agnostic loss, each time fixing the previously selected best weights. This sequential tuning strategy allowed us to identify the most effective configuration while minimizing computation cost. The best configuration obtained was $\gamma_t=1$, $\alpha=0.1$, and $\beta=0.5$. All experiments in the paper were conducted using this configuration to ensure consistency across results.

\begin{table}[h]
\centering
\renewcommand{\arraystretch}{1.3}
\begin{tabular}{c|ccc}
\hline
\textbf{Weights} & \multicolumn{3}{c}{\textbf{Accuracy}} \\
\cmidrule(lr){2-4} 
 & \textbf{$\mathcal{L}_{unsup}^t$} & \textbf{$\mathcal{L}_{aw}$} & \textbf{$\mathcal{L}_{agn}$} \\
\hline\hline
\textbf{0.01} & 38.6 & 40.7 & 40.9 \\
\textbf{0.05} & 40.5 & 40.7 & 43.8 \\
\textbf{0.1}* & 40.3 & \textbf{43.9} & 41.4 \\
\textbf{0.3}  & 39.9 & 39.9 & 40.1 \\
\textbf{0.5}* & 38.7 & 40.3 & \textbf{44.1} \\
\textbf{0.7}  & 39.9 & 39.9 & 39.9 \\
\textbf{0.9}  & 40.2 & 41.4 & 43.9 \\
\textbf{1}*   & \textbf{40.7} & 40.7 & 40.9 \\
\textbf{2}    & 40.6 & 39.6 & 41.9 \\
\textbf{5}    & 40.2 & 38.8 & 22.0 \\
\hline
\end{tabular}
\caption{Ablation study on loss weights. Each loss weight is varied while keeping the previously selected optimal weights fixed.}
\label{tab:loss_weight_ablation}
\end{table}

\begin{table*}[ht]
\centering
\renewcommand{\arraystretch}{1.3}
\begin{tabular}{l|cccccccccc|c|c}
\hline
$\tau_{ss}$ & Sub-1 & Sub-2 & Sub-3 & Sub-4 & Sub-5 & Sub-6 & Sub-7 & Sub-8 & Sub-9 & Sub-10 & Avg SS & Avg Acc \\
\hline\hline
0.00 & 77 & 77 & 77 & 77 & 77 & 77 & 77 & 77 & 77 & 77 & 77.0 & 37.66 \\
0.05 & 56 & 55 & 62 & 52 & 57 & 50 & 59 & 53 & 52 & 65 & 56.1 & 37.59 \\
0.10 & 53 & 50 & 57 & 52 & 53 & 48 & 57 & 52 & 49 & 63 & 53.4 & 38.02 \\
0.15 & 50 & 47 & 56 & 49 & 49 & 45 & 52 & 50 & 44 & 61 & 50.3 & 33.92 \\
0.20 & 48 & 44 & 50 & 47 & 45 & 43 & 48 & 44 & 40 & 54 & 46.3 & 36.51 \\
0.25 & 43 & 39 & 44 & 44 & 45 & 43 & 41 & 40 & 35 & 51 & 42.5 & 36.70 \\
0.30 & 39 & 35 & 40 & 37 & 43 & 41 & 37 & 38 & 33 & 49 & 39.2 & 38.90 \\
0.35 & 35 & 28 & 34 & 35 & 38 & 36 & 32 & 36 & 29 & 44 & 34.7 & 35.52 \\
0.40 & 31 & 22 & 28 & 32 & 35 & 33 & 30 & 28 & 27 & 39 & 30.5 & 37.27 \\
0.45 & 31 & 20 & 26 & 26 & 31 & 30 & 25 & 20 & 21 & 35 & 26.5 & 38.39 \\
0.50 & 26 & 16 & 24 & 26 & 28 & 26 & 21 & 16 & 18 & 34 & 23.5 & 38.69 \\
\textbf{0.55} & \textbf{20} & \textbf{23} & \textbf{23} & \textbf{24} & \textbf{22} & \textbf{23} & \textbf{16} & \textbf{13} & \textbf{15} & \textbf{30} & \textbf{20.9} & \textbf{39.89} \\
0.60 & 16 & 15 & 21 & 20 & 15 & 18 & 16 & 10 & 12 & 27 & 17.0 & 38.29 \\
0.65 & 12 & 13 & 20 & 19 & 15 & 16 & 11 & 11 & 10 & 22 & 14.9 & 37.62 \\
0.70 & 9 & 8 & 15 & 15 & 12 & 15 & 7 & 6 & 9 & 21 & 11.7 & 37.48 \\
0.75 & 7 & 6 & 13 & 12 & 10 & 13 & 5 & 5 & 8 & 20 & 9.9 & 37.42 \\
0.80 & 5 & 5 & 9 & 7 & 6 & 11 & 3 & 4 & 8 & 16 & 7.4 & 31.31 \\
0.85 & 4 & 4 & 6 & 7 & 5 & 9 & 3 & 3 & 5 & 13 & 5.9 & 30.27 \\
0.90 & 2 & 2 & 6 & 6 & 4 & 7 & 3 & 3 & 3 & 10 & 4.6 & 27.93 \\
0.95 & 2 & 2 & 4 & 3 & 3 & 3 & 2 & 2 & 3 & 5 & 2.9 & 28.53 \\
1.00 & 2 & 2 & 2 & 2 & 2 & 2 & 2 & 2 & 2 & 2 & 2.0 & 29.12 \\
\hline
\end{tabular}
\caption{Number of selected subjects (SS) for each target subject across different $\tau_{ss}$ thresholds. Lower $\tau_{ss}$ values indicate fewer selected sources. We also report the average number of selected subjects and the corresponding average accuracy.}
\label{tab:subject_selection}
\end{table*}

\subsection{Source Subject Selection Analysis}
Tab.~\ref{tab:subject_selection} shows the number of selected source subjects for each of the 10 target subjects for the Biovid dataset for various threshold ($\tau_{ss}$) values. As $\tau_{ss}$ increases, fewer source subjects are selected, reducing potential noise from irrelevant domains. Notably, the accuracy improves when transitioning from $\tau=0$ to moderate thresholds (e.g., $\tau_{ss}=0.45$ or $0.55$), demonstrating that selective inclusion of source data benefits generalization. The best average accuracy of 39.89\% is achieved with $\tau_{ss}=0.55$, using approximately 21 subjects on average. To assess the robustness of the source subject selection threshold ($\tau_{ss}$), we conducted experiments on the Behavioural Ambivalence/Hesitancy (BAH) \cite{gonzalez2025bah} dataset, which differs substantially from lab-controlled datasets (BioVid and Stress) as it consists of self-recorded, in-the-wild videos. Table~\ref{tab:ss_threshold_bah} shows results across thresholds from 0.45 to 0.65. The analysis reveals that accuracy remains stable in the 0.45–0.55 range, with the best performance achieved at $\tau_{ss}=0.55$. Higher thresholds reduce the number of selected sources and lead to performance degradation due to over-pruning. These findings confirm that $\tau_{ss}$ is not highly sensitive to fine-tuning and that a default value around 0.55 provides a reliable trade-off between accuracy and source diversity across datasets.

\subsection{Ablation on Target PLs Threshold ($\tau_{pl}$).} Tab.~\ref{tab:threshold_results}, and found that $\tau_{0}=0.95$ provided the best trade-off between label quality and sample quantity. Rather than fixing this value, we employ a confidence-annealing schedule, where $\tau_{pl}$ is decreased by $0.01$ every $N$ epochs, allowing more target samples to be included as the model stabilizes. This ensures early training benefits from high-confidence labels, while later stages incorporate a broader set of samples. To further mitigate the exclusion of useful but uncertain samples, we introduce a domain-agnostic loss that allows learning from low-confidence examples as well. Target ground-truth labels were only used here to evaluate pseudo-label quality.

\begin{table}[h]
\centering
\renewcommand{\arraystretch}{1.3}
\scriptsize
\begin{tabular}{c|ccccc|c|c}
\hline
{$\tau_{ss}$} & {Sub-1} & {Sub-2} & {Sub-3} & {Sub-4} & {Sub-5} & {Avg SS} & {Avg Acc} \\
\hline\hline
0.45 & 44 & 33 & 39 & 41 & 39 & 39.2 & 67.1 \\
0.50 & 42 & 31 & 36 & 40 & 36 & 37.0 & 68.2 \\
\textbf{0.55} & \textbf{41} & \textbf{30} & \textbf{35} & \textbf{38} & \textbf{34} & \textbf{35.6} & \textbf{68.9} \\
0.60 & 39 & 28 & 33 & 37 & 33 & 34.0 & 66.6 \\
0.65 & 38 & 27 & 30 & 36 & 30 & 32.2 & 62.5 \\
\hline
\end{tabular}
\caption{Impact of varying source subject selection threshold $\tau_{ss}$ on the BAH dataset.}
\label{tab:ss_threshold_bah}
\end{table}

\begin{table}[h]
\centering
\begin{tabular}{@{}lc@{}}
\toprule
\textbf{Threshold ($\tau_{pl}$)} & \textbf{Accuracy} \\ 
\midrule
0.50 & 40.2 \\
0.60 & 48.1 \\
0.70 & 44.7 \\
0.75 & 46.5 \\
0.80 & 46.2 \\
0.85 & 48.6 \\
0.90 & 46.7 \\
\textbf{0.95} & \textbf{52.0} \\ 
\bottomrule
\end{tabular}
\caption{Accuracy at different pseudo-label confidence thresholds ($\tau_{pl}$)}
\label{tab:threshold_results}
\end{table}


\begin{table}
\footnotesize
\centering
\renewcommand{\arraystretch}{1.4}
\begin{tabular}{ccc|cc}
\hline
\textbf{Exp-head} & \textbf{Id-head} & \textbf{Knife} & \textbf{Visual} & \textbf{Physio} \\ \hline \hline
\checkmark &  &  & 25.1 & 36.2 \\
\checkmark &  & \checkmark & 26.3 & 37.0 \\
\checkmark & \checkmark & \checkmark & \textbf{28.3} & \textbf{38.2} \\ \hline
\end{tabular}
\caption{Ablation on the disentanglement module for visual and physiological modalities for the Biovid dataset.}
\label{tab:distn_abl}
\end{table}

\subsection{Impact of Disentanglement}
\begin{figure}[t!]
\centering
\includegraphics[width=1.00\linewidth]{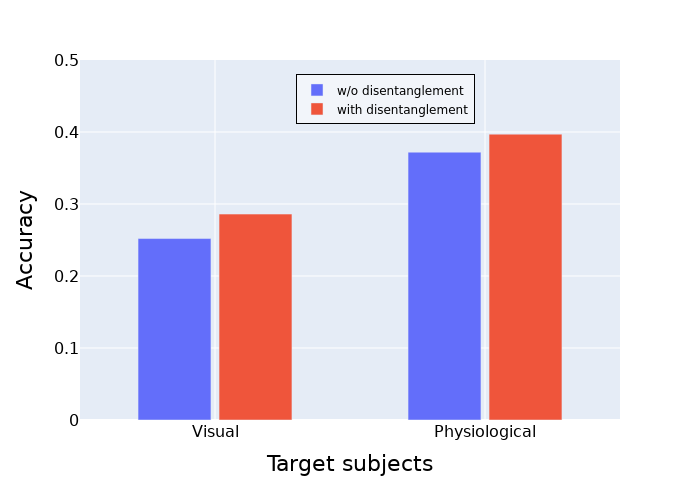}
\caption{Impact of adding disentanglement.}
\label{fig:ablation}      
\vspace{-10pt}
\end{figure}

conducted separately for visual and physiological modalities on the BioVid dataset, using 77 source and 10 target subjects. Results (Fig. \ref{fig:ablation}, Tab. \ref{tab:distn_abl}) show that incorporating the KNIFE-based disentanglement improves performance for both modalities. Without disentanglement, identity information interferes with expression classification, resulting in reduced accuracy. Adding the KNIFE-loss mitigates this, and introducing an identity head yields further improvements, confirming that explicitly removing identity bias enhances modality-specific expression features. 

\subsection{Computational Complexity} To analyze the time complexity, we evaluated all methods under identical conditions using a ResNet-18 backbone (11.7M parameters), a fixed batch size of 32, and the same GPU (NVIDIA A100-SXM4-40GB). The reported training time includes both source and target, and inference time with accuracy are summarized in Tab.~\ref{tab:timing_comparison}. The significant training time of CAN \cite{kang2020contrastive} is due to its design, which processes each source domain separately and performs class-wise contrastive learning between each source and the target domain within every batch. While manageable for benchmark datasets like Office-31 (3 domains), this becomes prohibitively expensive when scaling to our subject-based adaptation setting, where the number of source domains averages over 30, resulting in 19,625 ms of target training time. In contrast, our method retrieves class-wise samples from all source domains and from the target domain, then performs a single contrastive operation per batch, significantly reducing overhead. It completes target training in just 426 ms (46$\times$ faster than CAN) while improving accuracy from 34.6\% to 43.8\%. Although CMSDA \cite{scalbert2021multi} and sub-based \cite{zeeshan2024subject} adaptation methods are more efficient (271 ms and 234 ms), they fail to yield improvements in accuracy. Results show that our method offers the best trade-off between efficiency and accuracy, with inference time similar to other baselines. 

\begin{table}[h]
\centering
\renewcommand{\arraystretch}{1.3}
\begin{tabular}{c|cccc}
\hline
\textbf{Method} & \textbf{Src Train} & \textbf{Trg Train} & \textbf{Inf} & \textbf{Acc} \\
\hline\hline
CAN~\cite{kang2020contrastive}       & 59     & 19,625  & 8.03  & 34.6 \\
Sub-based~\cite{zeeshan2024subject} & 59     & 234     & 7.36  & 34.6 \\
CMSDA~\cite{scalbert2021multi}     & --     & 271     & 4.07  & 36.8 \\
Ours (all srcs) & 59 & 1895 & 7.82 & 39.3 \\
Ours (co-train)      & \textbf{59}     & \textbf{426}     & \textbf{7.64}  & \textbf{43.8} \\
\bottomrule
\end{tabular}
\caption{Training and inference time (per batch) of our proposed and baseline methods. All times are shown in ms.}
\label{tab:timing_comparison}
\end{table}

\begin{table}[h]
\centering
\footnotesize
\renewcommand{\arraystretch}{1.3}
\begin{tabular}{cccc|c}
\hline
$\mathcal{L}^s$ & $\mathcal{L}^t$ & $\mathcal{L}_{aw}$ & $\mathcal{L}_{agn}$ & \textbf{Avg Acc} \\
\hline\hline
\checkmark &  &  &  & 32.1 \\
\checkmark & \checkmark &  &  & 36.5 \\
\checkmark & \checkmark & \checkmark & & 38.0 \\ 
\checkmark & \checkmark &  & \checkmark  & 37.2 \\
\checkmark & \checkmark & \checkmark & \checkmark & \textbf{40.7} \\
\hline
\end{tabular}
\caption{Impact of individual loss components under equally weighted hyperparameter.}
\label{tab:ab_eq_we_loss_and_tau}
\vspace{-2em}  
\end{table}

\begin{table*}
\scriptsize
\renewcommand{\arraystretch}{1.4}
\centering
\begin{tabular}{c|c|ccccccccccc}
\hline
\multirow{2}{*}{\textbf{Settings}} &
\multirow{2}{*}{\textbf{Methods}} &
\multicolumn{11}{c}{\textbf{Target Subjects}} \\ \cline{3-13}
& & \textbf{Sub-1} & \textbf{Sub-2} & \textbf{Sub-3} & \textbf{Sub-4} & \textbf{Sub-5} & \textbf{Sub-6} & \textbf{Sub-7} & \textbf{Sub-8} & \textbf{Sub-9} & \textbf{Sub-10} & \textbf{Avg F1} \\ \hline \hline

Lower Bound &
\begin{tabular}[c]{@{}c@{}}Visual-only\\Physio-only\\Fusion\end{tabular} &
\begin{tabular}[c]{@{}c@{}}63.1\\36.6\\57.3\end{tabular} &
\begin{tabular}[c]{@{}c@{}}24.6\\45.6\\32.1\end{tabular} &
\begin{tabular}[c]{@{}c@{}}61.2\\37.7\\50.0\end{tabular} &
\begin{tabular}[c]{@{}c@{}}37.8\\22.3\\36.6\end{tabular} &
\begin{tabular}[c]{@{}c@{}}58.3\\37.5\\51.5\end{tabular} &
\begin{tabular}[c]{@{}c@{}}50.8\\40.8\\63.4\end{tabular} &
\begin{tabular}[c]{@{}c@{}}43.2\\45.6\\68.4\end{tabular} &
\begin{tabular}[c]{@{}c@{}}36.6\\33.8\\39.5\end{tabular} &
\begin{tabular}[c]{@{}c@{}}46.9\\25.6\\52.6\end{tabular} &
\begin{tabular}[c]{@{}c@{}}54.8\\22.7\\{53.4}\end{tabular} &
\begin{tabular}[c]{@{}c@{}}47.7\\34.8\\50.5\end{tabular} \\ \hline

\begin{tabular}[c]{@{}c@{}}MM-UDA\\[-4pt](Blending)\end{tabular} &
\begin{tabular}[c]{@{}c@{}}DANN \cite{ganin2016domain}\\CDAN \cite{long2018conditional}\\MMD \cite{sejdinovic2013equivalence}\\ \oursuda\end{tabular} &
\begin{tabular}[c]{@{}c@{}}61.2\\57.1\\57.1\\37.6\end{tabular} &
\begin{tabular}[c]{@{}c@{}}45.3\\43.2\\54.0\\49.3\end{tabular} &
\begin{tabular}[c]{@{}c@{}}55.6\\64.3\\61.3\\74.1\end{tabular} &
\begin{tabular}[c]{@{}c@{}}51.3\\46.6\\40.3\\{86.2}\end{tabular} &
\begin{tabular}[c]{@{}c@{}}53.6\\48.3\\45.6\\55.7\end{tabular} &
\begin{tabular}[c]{@{}c@{}}56.6\\58.3\\55.6\\61.5\end{tabular} &
\begin{tabular}[c]{@{}c@{}}56.6\\59.6\\55.6\\64.8\end{tabular} &
\begin{tabular}[c]{@{}c@{}}47.6\\45.6\\{55.6}\\53.4\end{tabular} &
\begin{tabular}[c]{@{}c@{}}62.3\\62.2\\55.4\\{74.4}\end{tabular} &
\begin{tabular}[c]{@{}c@{}}38.6\\39.6\\49.3\\50.0\end{tabular} &
\begin{tabular}[c]{@{}c@{}}52.9\\52.5\\53.0\\60.7\end{tabular} \\ \hline

MM-MSDA &
\begin{tabular}[c]{@{}c@{}}CAN \cite{kang2020contrastive}\\Sub-based\textsubscript{top-k} \cite{zeeshan2024subject}\\ \oursmsda \end{tabular} &
\begin{tabular}[c]{@{}c@{}}65.3\\68.9\\ {76.1}\end{tabular} &
\begin{tabular}[c]{@{}c@{}}45.4\\45.3\\{59.6}\end{tabular} &
\begin{tabular}[c]{@{}c@{}}54.6\\{74.2}\\69.3\end{tabular} &
\begin{tabular}[c]{@{}c@{}}42.3\\46.6\\53.3\end{tabular} &
\begin{tabular}[c]{@{}c@{}}51.2\\55.2\\{58.6}\end{tabular} &
\begin{tabular}[c]{@{}c@{}}58.6\\62.6\\{67.2}\end{tabular} &
\begin{tabular}[c]{@{}c@{}}66.6\\56.6\\{75.2}\end{tabular} &
\begin{tabular}[c]{@{}c@{}}54.2\\39.3\\48.3\end{tabular} &
\begin{tabular}[c]{@{}c@{}}59.3\\52.2\\{74.6}\end{tabular} &
\begin{tabular}[c]{@{}c@{}}34.6\\39.9\\40.2\end{tabular} &
\begin{tabular}[c]{@{}c@{}}53.2\\54.1\\\textbf{62.2}\end{tabular} \\ \hline

Upper Bound & Fine-tuning &
86.3 & 43.8 & 94.4 & 97.7 & 79.6 & 75.6 & 45.6 & 93.0 & 91.0 & 97.0 & 80.4 \\ \hline
\end{tabular}
\caption{Subject-wise F1 scores of StressID dataset across 10 target subjects.}
\label{tab:stressid_f1}
\end{table*}

\begin{table*}
\scriptsize
\renewcommand{\arraystretch}{1.4}
\centering
\begin{tabular}{c|c|ccccccccccc}
\hline
\multirow{2}{*}{\textbf{Settings}} &
\multirow{2}{*}{\textbf{Methods}} &
\multicolumn{11}{c}{\textbf{Target Subjects}} \\ \cline{3-13}
& & \textbf{Sub-1} & \textbf{Sub-2} & \textbf{Sub-3} & \textbf{Sub-4} & \textbf{Sub-5} & \textbf{Sub-6} & \textbf{Sub-7} & \textbf{Sub-8} & \textbf{Sub-9} & \textbf{Sub-10} & \textbf{Avg F1} \\ \hline \hline

Lower Bound &
\begin{tabular}[c]{@{}c@{}}Visual-only\\Physio-only\\Fusion\end{tabular} &
\begin{tabular}[c]{@{}c@{}}29.8\\38.1\\41.2\end{tabular} &
\begin{tabular}[c]{@{}c@{}}26.1\\32.5\\27.2\end{tabular} &
\begin{tabular}[c]{@{}c@{}}9.1\\39.1\\24.9\end{tabular} &
\begin{tabular}[c]{@{}c@{}}28.1\\15.4\\32.2\end{tabular} &
\begin{tabular}[c]{@{}c@{}}19.3\\10.4\\29.5\end{tabular} &
\begin{tabular}[c]{@{}c@{}}21.6\\29.0\\25.8\end{tabular} &
\begin{tabular}[c]{@{}c@{}}25.7\\14.3\\34.7\end{tabular} &
\begin{tabular}[c]{@{}c@{}}15.3\\34.6\\24.3\end{tabular} &
\begin{tabular}[c]{@{}c@{}}16.5\\28.7\\22.3\end{tabular} &
\begin{tabular}[c]{@{}c@{}}16.1\\22.0\\18.9\end{tabular} &
\begin{tabular}[c]{@{}c@{}}20.8\\26.4\\28.1\end{tabular} \\ \hline

\begin{tabular}[c]{@{}c@{}}MM-UDA\\[-4pt](Blending)\end{tabular} &
\begin{tabular}[c]{@{}c@{}}DANN \cite{ganin2016domain}\\CDAN \cite{long2018conditional}\\MMD \cite{sejdinovic2013equivalence}\\ \oursuda \end{tabular} &
\begin{tabular}[c]{@{}c@{}}32.5\\28.5\\28.3\\35.6\end{tabular} &
\begin{tabular}[c]{@{}c@{}}25.6\\24.3\\20.7\\21.7\end{tabular} &
\begin{tabular}[c]{@{}c@{}}29.5\\25.5\\29.6\\33.3\end{tabular} &
\begin{tabular}[c]{@{}c@{}}33.6\\29.2\\32.4\\{36.2}\end{tabular} &
\begin{tabular}[c]{@{}c@{}}28.3\\25.2\\28.4\\28.7\end{tabular} &
\begin{tabular}[c]{@{}c@{}}25.1\\24.2\\29.5\\16.0\end{tabular} &
\begin{tabular}[c]{@{}c@{}}29.6\\28.6\\32.1\\31.2\end{tabular} &
\begin{tabular}[c]{@{}c@{}}24.2\\23.4\\28.6\\29.4\end{tabular} &
\begin{tabular}[c]{@{}c@{}}25.2\\23.1\\25.4\\35.4\end{tabular} &
\begin{tabular}[c]{@{}c@{}}20.2\\20.5\\26.5\\17.6\end{tabular} &
\begin{tabular}[c]{@{}c@{}}27.4\\25.3\\28.2\\28.5\end{tabular} \\ \hline

MM-MSDA &
\begin{tabular}[c]{@{}c@{}}CAN \cite{kang2020contrastive}\\Sub-based\textsubscript{top-k} \cite{zeeshan2024subject}\\CMSDA \cite{scalbert2021multi}\\ \oursmsda \end{tabular} &
\begin{tabular}[c]{@{}c@{}}34.6\\34.6\\31.2\\{41.6}\end{tabular} &
\begin{tabular}[c]{@{}c@{}}30.3\\{37.6}\\32.5\\33.2\end{tabular} &
\begin{tabular}[c]{@{}c@{}}22.3\\22.4\\21.2\\{49.6}\end{tabular} &
\begin{tabular}[c]{@{}c@{}}24.6\\20.5\\31.2\\27.8\end{tabular} &
\begin{tabular}[c]{@{}c@{}}27.9\\{34.5}\\32.3\\25.6\end{tabular} &
\begin{tabular}[c]{@{}c@{}}26.3\\24.6\\30.1\\{36.5}\end{tabular} &
\begin{tabular}[c]{@{}c@{}}28.6\\{36.8}\\30.2\\29.4\end{tabular} &
\begin{tabular}[c]{@{}c@{}}26.6\\24.3\\34.1\\{35.4}\end{tabular} &
\begin{tabular}[c]{@{}c@{}}33.2\\30.6\\38.9\\{49.0}\end{tabular} &
\begin{tabular}[c]{@{}c@{}}18.1\\17.6\\18.5\\{29.2}\end{tabular} &
\begin{tabular}[c]{@{}c@{}}27.3\\28.4\\30.0\\\textbf{35.7}\end{tabular} \\ \hline

Upper Bound & Fine-tuning &
81.2 & 74.3 & 67.6 & 68.2 & 70.3 & 61.4 & 74.2 & 70.6 & 74.0 & 49.5 & 69.1 \\ \hline
\end{tabular}
\caption{Subject-wise F1 scores of BioVid dataset on 10 target subjects.}
\label{tab:biovid_f1}
\end{table*}

\begin{table*}[h]
\centering
\scriptsize
\renewcommand{\arraystretch}{1.3}
\begin{tabular}{c|c|ccccc|c}
\hline
\textbf{Settings} & \textbf{Methods} & \textbf{Sub-1} & \textbf{Sub-2} & \textbf{Sub-3} & \textbf{Sub-4} & \textbf{Sub-5} & \textbf{Avg F1} \\ \hline \hline

\multirow{3}{*}{Lower Bound} & Visual-only & 25.2 & 46.2 & 35.2 & 41.3 & 43.2 & 38.2 \\
 & Audio-only & 27.4 & 34.4 & 34.1 & 40.3 & 44.1 & 36.1 \\
 & Fusion & 26.3 & 44.5 & 34.7 & 41.9 & 44.6 & 38.4 \\ \hline

\multirow{2}{*}{MM-UDA} & MMD~\cite{sejdinovic2013equivalence} & 28.2 & 44.5 & 39.3 & 38.7 & 43.2 & 38.8 \\
 & \oursuda & 26.1 & 46.1 & 35.6 & 42.2 & 43.5 & 38.7 \\ \hline

\multirow{2}{*}{MM-MSDA} & Sub-based\textsubscript{top-k}~\cite{zeeshan2024subject} & 24.6 & 45.4 & 36.8 & 42.7 & 44.9 & 38.9 \\
 & \oursmsda & \textbf{28.2} & \textbf{48.6} & \textbf{39.4} & \textbf{44.2} & \textbf{45.4} & \textbf{41.2} \\ \hline

\end{tabular}
\caption{Subject-wise F1 scores on the BAH dataset across five target subjects. MuSACo MSDA consistently outperforms baselines and other UDA/MSDA methods.}
\label{tab:bah_f1}
\end{table*}

\subsection{Impact of Difference Loss Components with Equal Weights}
In addition to the tuned-weight ablation reported in the main paper, we also conducted a diagnostic experiment where all loss weights were fixed to 1. This isolates the relative impact of each component without any hyperparameter tuning. Table~\ref{tab:ab_eq_we_loss_and_tau} shows that the full model under equal weights reached 40.7\%, which is lower than the tuned configuration (43.8\%). Importantly, the incremental trend remains consistent across both setups: $\mathcal{L}^t$, $\mathcal{L}_{aw}$, and $\mathcal{L}_{agn}$ each contribute positively to performance. This confirms that the effectiveness of MuSACo does not hinge on carefully tuned weights, but rather that appropriate weighting further amplifies the gains. Together, these results underline the robustness and general applicability of the proposed loss formulation.

\subsection{Target Pseudo-label Progression During Training}
\begin{figure}[t!]
\centering
\includegraphics[width=1.00\linewidth]{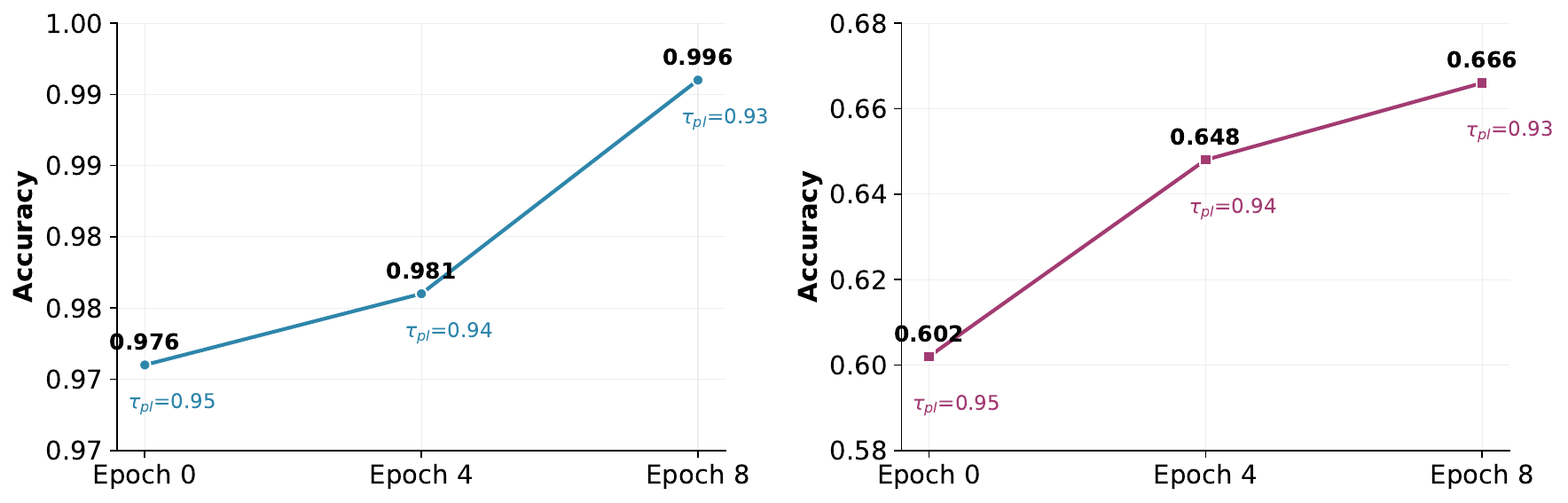}
\caption{Target pseudo-labels accuracy during training.}
\label{fig:pl_progress}      
\vspace{-10pt}
\end{figure}

We evaluate the reliability of co-training by tracking pseudo-label (PL) accuracy during training. Results are reported for two challenging StressID subjects: ctzy (Sub-4) and x1q3 (Sub-10), which exhibit high label imbalance. To stabilize PL quality, we adopt a confidence-annealing schedule, initializing $\tau_{pl}=0.95$ and gradually reducing it by 0.01 every $N$ epochs (e.g., 0.94 at Ep-4, 0.93 at Ep-8). As shown in Fig.~\ref{fig:pl_progress}, PL accuracy steadily improves across epochs, demonstrating the effectiveness of iterative refinement under this schedule. For Sub-4, PL accuracy improves from 0.976 at initialization to 0.996 after 8 epochs, while for Sub-10, which is highly imbalanced, accuracy increases from 0.602 to 0.666. These trends indicate that co-training with confidence annealing reliably enhances PL quality, even under challenging conditions.

\begin{table*}
\scriptsize
\renewcommand{\arraystretch}{1.4}
\centering
\begin{tabular}{c|c|ccccccccccc}
\hline
\multirow{2}{*}{\textbf{Settings}} &
\multirow{2}{*}{\textbf{Methods}} &
\multicolumn{11}{c}{\textbf{Target Subjects}} \\ \cline{3-13}
& & \textbf{Sub-1} & \textbf{Sub-2} & \textbf{Sub-3} & \textbf{Sub-4} & \textbf{Sub-5} & \textbf{Sub-6} & \textbf{Sub-7} & \textbf{Sub-8} & \textbf{Sub-9} & \textbf{Sub-10} & \textbf{Avg Acc} \\ \hline \hline

Lower Bound &
\begin{tabular}[c]{@{}c@{}}Visual-only\\Physio-only\\Fusion\end{tabular} &
\begin{tabular}[c]{@{}c@{}}37.8\\41.7\\36.0\end{tabular} &
\begin{tabular}[c]{@{}c@{}}27.7\\30.1\\28.0\end{tabular} &
\begin{tabular}[c]{@{}c@{}}23.3\\30.5\\26.0\end{tabular} &
\begin{tabular}[c]{@{}c@{}}22.6\\33.3\\35.0\end{tabular} &
\begin{tabular}[c]{@{}c@{}}27.5\\36.7\\35.0\end{tabular} &
\begin{tabular}[c]{@{}c@{}}21.6\\30.9\\15.0\end{tabular} &
\begin{tabular}[c]{@{}c@{}}37.0\\34.7\\43.0\end{tabular} &
\begin{tabular}[c]{@{}c@{}}20.0\\38.0\\40.0\end{tabular} &
\begin{tabular}[c]{@{}c@{}}22.0\\37.3\\35.0\end{tabular} &
\begin{tabular}[c]{@{}c@{}}24.0\\28.0\\28.0\end{tabular} &
\begin{tabular}[c]{@{}c@{}}26.4\\34.1\\32.1\end{tabular} \\ \hline

MM-UDA &
\begin{tabular}[c]{@{}c@{}}MMD~\cite{sejdinovic2013equivalence}\\ \oursuda\end{tabular} &
\begin{tabular}[c]{@{}c@{}}40.3\\39.0\end{tabular} &
\begin{tabular}[c]{@{}c@{}}31.2\\35.0\end{tabular} &
\begin{tabular}[c]{@{}c@{}}32.3\\34.0\end{tabular} &
\begin{tabular}[c]{@{}c@{}}26.3\\32.0\end{tabular} &
\begin{tabular}[c]{@{}c@{}}33.3\\29.0\end{tabular} &
\begin{tabular}[c]{@{}c@{}}24.1\\19.0\end{tabular} &
\begin{tabular}[c]{@{}c@{}}34.3\\38.9\end{tabular} &
\begin{tabular}[c]{@{}c@{}}39.5\\37.0\end{tabular} &
\begin{tabular}[c]{@{}c@{}}37.3\\31.0\end{tabular} &
\begin{tabular}[c]{@{}c@{}}20.0\\29.9\end{tabular} &
\begin{tabular}[c]{@{}c@{}}31.9\\32.5\end{tabular} \\ \hline

MM-MSDA &
\begin{tabular}[c]{@{}c@{}}Sub-based\textsubscript{top-k}~\cite{zeeshan2024subject}\\ \oursmsda\end{tabular} &
\begin{tabular}[c]{@{}c@{}}45.0\\42.0\end{tabular} &
\begin{tabular}[c]{@{}c@{}}35.0\\38.0\end{tabular} &
\begin{tabular}[c]{@{}c@{}}39.0\\51.0\end{tabular} &
\begin{tabular}[c]{@{}c@{}}37.0\\30.0\end{tabular} &
\begin{tabular}[c]{@{}c@{}}34.0\\28.0\end{tabular} &
\begin{tabular}[c]{@{}c@{}}32.0\\39.0\end{tabular} &
\begin{tabular}[c]{@{}c@{}}42.0\\43.0\end{tabular} &
\begin{tabular}[c]{@{}c@{}}36.0\\35.0\end{tabular} &
\begin{tabular}[c]{@{}c@{}}27.0\\43.0\end{tabular} &
\begin{tabular}[c]{@{}c@{}}30.0\\33.0\end{tabular} &
\begin{tabular}[c]{@{}c@{}}35.7\\\textbf{38.2}\end{tabular} \\ \hline

\end{tabular}
\caption{Subject-wise accuracy of the BioVid dataset using the ViT backbone across 10 target subjects.}
\label{tab:biovid_vit_acc}
\end{table*}

\begin{table*}
\scriptsize
\renewcommand{\arraystretch}{1.4}
\centering
\begin{tabular}{c|c|ccccccccccc}
\hline
\multirow{2}{*}{\textbf{Settings}} &
\multirow{2}{*}{\textbf{Methods}} &
\multicolumn{11}{c}{\textbf{Target Subjects}} \\ \cline{3-13}
& & \textbf{Sub-1} & \textbf{Sub-2} & \textbf{Sub-3} & \textbf{Sub-4} & \textbf{Sub-5} & \textbf{Sub-6} & \textbf{Sub-7} & \textbf{Sub-8} & \textbf{Sub-9} & \textbf{Sub-10} & \textbf{Avg F1} \\ \hline \hline

Lower Bound &
\begin{tabular}[c]{@{}c@{}}Visual-only\\Physio-only\\Fusion\end{tabular} &
\begin{tabular}[c]{@{}c@{}}35.6\\31.0\\23.9\end{tabular} &
\begin{tabular}[c]{@{}c@{}}27.5\\21.9\\27.1\end{tabular} &
\begin{tabular}[c]{@{}c@{}}22.5\\23.1\\22.8\end{tabular} &
\begin{tabular}[c]{@{}c@{}}12.5\\24.5\\22.6\end{tabular} &
\begin{tabular}[c]{@{}c@{}}19.7\\27.9\\33.5\end{tabular} &
\begin{tabular}[c]{@{}c@{}}19.6\\23.6\\9.5\end{tabular} &
\begin{tabular}[c]{@{}c@{}}35.3\\23.9\\32.9\end{tabular} &
\begin{tabular}[c]{@{}c@{}}15.6\\27.3\\34.6\end{tabular} &
\begin{tabular}[c]{@{}c@{}}14.9\\28.2\\29.1\end{tabular} &
\begin{tabular}[c]{@{}c@{}}23.2\\19.1\\26.1\end{tabular} &
\begin{tabular}[c]{@{}c@{}}22.6\\25.0\\26.2\end{tabular} \\ \hline

MM-UDA &
\begin{tabular}[c]{@{}c@{}}MMD~\cite{sejdinovic2013equivalence}\\ \oursuda\end{tabular} &
\begin{tabular}[c]{@{}c@{}}31.7\\33.6\end{tabular} &
\begin{tabular}[c]{@{}c@{}}33.1\\34.6\end{tabular} &
\begin{tabular}[c]{@{}c@{}}28.6\\25.9\end{tabular} &
\begin{tabular}[c]{@{}c@{}}20.5\\26.9\end{tabular} &
\begin{tabular}[c]{@{}c@{}}33.8\\25.9\end{tabular} &
\begin{tabular}[c]{@{}c@{}}23.5\\18.4\end{tabular} &
\begin{tabular}[c]{@{}c@{}}37.7\\38.4\end{tabular} &
\begin{tabular}[c]{@{}c@{}}35.6\\33.8\end{tabular} &
\begin{tabular}[c]{@{}c@{}}29.3\\31.3\end{tabular} &
\begin{tabular}[c]{@{}c@{}}18.1\\29.8\end{tabular} &
\begin{tabular}[c]{@{}c@{}}29.2\\29.9\end{tabular} \\ \hline

MM-MSDA &
\begin{tabular}[c]{@{}c@{}}Sub-based\textsubscript{top-k}~\cite{zeeshan2024subject}\\  \oursmsda\end{tabular} &
\begin{tabular}[c]{@{}c@{}}36.9\\33.9\end{tabular} &
\begin{tabular}[c]{@{}c@{}}27.2\\27.6\end{tabular} &
\begin{tabular}[c]{@{}c@{}}34.6\\38.7\end{tabular} &
\begin{tabular}[c]{@{}c@{}}32.9\\26.4\end{tabular} &
\begin{tabular}[c]{@{}c@{}}20.6\\24.9\end{tabular} &
\begin{tabular}[c]{@{}c@{}}27.4\\33.9\end{tabular} &
\begin{tabular}[c]{@{}c@{}}34.5\\41.5\end{tabular} &
\begin{tabular}[c]{@{}c@{}}28.0\\30.1\end{tabular} &
\begin{tabular}[c]{@{}c@{}}13.1\\37.6\end{tabular} &
\begin{tabular}[c]{@{}c@{}}22.6\\29.8\end{tabular} &
\begin{tabular}[c]{@{}c@{}}27.8\\ \textbf{32.4}\end{tabular} \\ \hline

\end{tabular}
\caption{Subject-wise F1 scores of the BioVid dataset using the ViT backbone across 10 target subjects.}
\label{tab:biovid_vit_f1}
\end{table*}

\begin{table*}
\scriptsize
\renewcommand{\arraystretch}{1.4}
\centering
\begin{tabular}{c|c|ccccccccccc}
\hline
\multirow{2}{*}{\textbf{Settings}} &
\multirow{2}{*}{\textbf{Methods}} &
\multicolumn{11}{c}{\textbf{Target Subjects}} \\ \cline{3-13}
& & \textbf{Sub-1} & \textbf{Sub-2} & \textbf{Sub-3} & \textbf{Sub-4} & \textbf{Sub-5} & \textbf{Sub-6} & \textbf{Sub-7} & \textbf{Sub-8} & \textbf{Sub-9} & \textbf{Sub-10} & \textbf{Avg Acc} \\ \hline \hline

Lower Bound &
\begin{tabular}[c]{@{}c@{}}Visual-only\\Physio-only\\Fusion\end{tabular} &
\begin{tabular}[c]{@{}c@{}}57.0\\57.8\\48.3\end{tabular} &
\begin{tabular}[c]{@{}c@{}}18.3\\84.1\\14.0\end{tabular} &
\begin{tabular}[c]{@{}c@{}}41.7\\60.7\\49.3\end{tabular} &
\begin{tabular}[c]{@{}c@{}}68.3\\28.8\\78.3\end{tabular} &
\begin{tabular}[c]{@{}c@{}}62.4\\60.1\\66.0\end{tabular} &
\begin{tabular}[c]{@{}c@{}}47.5\\69.0\\49.2\end{tabular} &
\begin{tabular}[c]{@{}c@{}}84.2\\84.1\\74.0\end{tabular} &
\begin{tabular}[c]{@{}c@{}}47.0\\51.2\\51.3\end{tabular} &
\begin{tabular}[c]{@{}c@{}}65.5\\34.4\\64.2\end{tabular} &
\begin{tabular}[c]{@{}c@{}}69.7\\29.4\\73.5\end{tabular} &
\begin{tabular}[c]{@{}c@{}}56.2\\56.0\\56.8\end{tabular} \\ \hline

MM-UDA &
\begin{tabular}[c]{@{}c@{}}MMD~\cite{sejdinovic2013equivalence}\\ \oursuda\end{tabular} &
\begin{tabular}[c]{@{}c@{}}53.0\\64.0\end{tabular} &
\begin{tabular}[c]{@{}c@{}}23.0\\20.0\end{tabular} &
\begin{tabular}[c]{@{}c@{}}36.4\\38.0\end{tabular} &
\begin{tabular}[c]{@{}c@{}}82.0\\79.0\end{tabular} &
\begin{tabular}[c]{@{}c@{}}69.0\\65.0\end{tabular} &
\begin{tabular}[c]{@{}c@{}}47.6\\50.0\end{tabular} &
\begin{tabular}[c]{@{}c@{}}78.0\\82.0\end{tabular} &
\begin{tabular}[c]{@{}c@{}}46.6\\50.0\end{tabular} &
\begin{tabular}[c]{@{}c@{}}77.0\\66.0\end{tabular} &
\begin{tabular}[c]{@{}c@{}}66.0\\68.0\end{tabular} &
\begin{tabular}[c]{@{}c@{}}57.9\\58.2\end{tabular} \\ \hline

MM-MSDA &
\begin{tabular}[c]{@{}c@{}}Sub-based\textsubscript{top-k}~\cite{zeeshan2024subject}\\ \oursmsda\end{tabular} &
\begin{tabular}[c]{@{}c@{}}58.0\\38.0\end{tabular} &
\begin{tabular}[c]{@{}c@{}}15.0\\38.0\end{tabular} &
\begin{tabular}[c]{@{}c@{}}50.0\\87.0\end{tabular} &
\begin{tabular}[c]{@{}c@{}}72.0\\100.0\end{tabular} &
\begin{tabular}[c]{@{}c@{}}62.0\\67.0\end{tabular} &
\begin{tabular}[c]{@{}c@{}}82.0\\67.0\end{tabular} &
\begin{tabular}[c]{@{}c@{}}94.0\\85.0\end{tabular} &
\begin{tabular}[c]{@{}c@{}}82.0\\82.0\end{tabular} &
\begin{tabular}[c]{@{}c@{}}66.0\\66.0\end{tabular} &
\begin{tabular}[c]{@{}c@{}}67.0\\68.0\end{tabular} &
\begin{tabular}[c]{@{}c@{}}64.8\\ \textbf{69.8}\end{tabular} \\ \hline

\end{tabular}
\caption{Subject-wise accuracy of the StressID dataset across 10 target subjects using the ViT backbone.}
\label{tab:stressid_vit_acc}
\end{table*}

\begin{table*}
\scriptsize
\renewcommand{\arraystretch}{1.4}
\centering
\begin{tabular}{c|c|ccccccccccc}
\hline
\multirow{2}{*}{\textbf{Settings}} &
\multirow{2}{*}{\textbf{Methods}} &
\multicolumn{11}{c}{\textbf{Target Subjects}} \\ \cline{3-13}
& & \textbf{Sub-1} & \textbf{Sub-2} & \textbf{Sub-3} & \textbf{Sub-4} & \textbf{Sub-5} &
\textbf{Sub-6} & \textbf{Sub-7} & \textbf{Sub-8} & \textbf{Sub-9} & \textbf{Sub-10} & \textbf{Avg F1} \\ \hline \hline

Lower Bound &
\begin{tabular}[c]{@{}c@{}}Visual-only\\Physio-only\\Fusion\end{tabular} &
\begin{tabular}[c]{@{}c@{}}55.2\\36.6\\47.0\end{tabular} &
\begin{tabular}[c]{@{}c@{}}17.3\\45.7\\13.0\end{tabular} &
\begin{tabular}[c]{@{}c@{}}32.5\\37.8\\36.4\end{tabular} &
\begin{tabular}[c]{@{}c@{}}67.9\\22.4\\77.4\end{tabular} &
\begin{tabular}[c]{@{}c@{}}46.1\\37.5\\53.4\end{tabular} &
\begin{tabular}[c]{@{}c@{}}46.3\\40.8\\47.1\end{tabular} &
\begin{tabular}[c]{@{}c@{}}76.4\\45.7\\64.2\end{tabular} &
\begin{tabular}[c]{@{}c@{}}32.9\\33.9\\42.8\end{tabular} &
\begin{tabular}[c]{@{}c@{}}39.6\\25.6\\39.6\end{tabular} &
\begin{tabular}[c]{@{}c@{}}41.1\\22.7\\42.3\end{tabular} &
\begin{tabular}[c]{@{}c@{}}45.5\\34.9\\46.3\end{tabular} \\ \hline

MM-UDA &
\begin{tabular}[c]{@{}c@{}}MMD~\cite{sejdinovic2013equivalence}\\ \oursuda\end{tabular} &
\begin{tabular}[c]{@{}c@{}}51.6\\62.6\end{tabular} &
\begin{tabular}[c]{@{}c@{}}22.6\\19.3\end{tabular} &
\begin{tabular}[c]{@{}c@{}}27.5\\28.5\end{tabular} &
\begin{tabular}[c]{@{}c@{}}72.6\\77.4\end{tabular} &
\begin{tabular}[c]{@{}c@{}}55.2\\49.3\end{tabular} &
\begin{tabular}[c]{@{}c@{}}42.1\\48.5\end{tabular} &
\begin{tabular}[c]{@{}c@{}}72.5\\72.8\end{tabular} &
\begin{tabular}[c]{@{}c@{}}44.6\\34.8\end{tabular} &
\begin{tabular}[c]{@{}c@{}}44.0\\39.5\end{tabular} &
\begin{tabular}[c]{@{}c@{}}39.5\\40.2\end{tabular} &
\begin{tabular}[c]{@{}c@{}}47.2\\47.3\end{tabular} \\ \hline

MM-MSDA &
\begin{tabular}[c]{@{}c@{}}Sub-based\textsubscript{top-k}~\cite{zeeshan2024subject}\\ \oursmsda\end{tabular} &
\begin{tabular}[c]{@{}c@{}}56.0\\28.5\end{tabular} &
\begin{tabular}[c]{@{}c@{}}13.8\\36.4\end{tabular} &
\begin{tabular}[c]{@{}c@{}}46.5\\85.2\end{tabular} &
\begin{tabular}[c]{@{}c@{}}71.0\\98.8\end{tabular} &
\begin{tabular}[c]{@{}c@{}}38.0\\66.3\end{tabular} &
\begin{tabular}[c]{@{}c@{}}80.1\\66.2\end{tabular} &
\begin{tabular}[c]{@{}c@{}}87.3\\45.7\end{tabular} &
\begin{tabular}[c]{@{}c@{}}81.2\\81.2\end{tabular} &
\begin{tabular}[c]{@{}c@{}}39.5\\39.5\end{tabular} &
\begin{tabular}[c]{@{}c@{}}39.9\\40.2\end{tabular} &
\begin{tabular}[c]{@{}c@{}}55.3\\ \textbf{58.8}\end{tabular} \\ \hline

\end{tabular}
\caption{Subject-wise F1 scores of the StressID dataset across 10 target subjects using the ViT backbone.}
\label{tab:stressid_vit_f1}
\end{table*}

\section{Subject-wise F1 Scores}

\subsection{Stress-ID}
Table \ref{tab:stressid_f1} presents subject-wise F1 scores on the StressID dataset. While the dataset has a binary label distribution (stress vs. no-stress), clear variability emerges across subjects. For example, lower-bound fusion consistently yields higher F1 than either modality alone, showing the complementary nature of visual and physiological signals. However, without adaptation, predictions remain unbalanced across subjects, with several experiencing sharp drops (e.g., Sub-2, Sub-4). UDA methods improve F1 moderately on some subjects, but results remain inconsistent. In contrast, \oursmsda achieves the best overall performance, boosting recall on harder subjects (e.g., Sub-9) and producing the highest average F1 across all targets. These findings underscore the importance of F1 as a complementary metric to accuracy, particularly for stress recognition, where subject-specific imbalance strongly affects model robustness.

\subsection{Biovid}
Tab.~\ref{tab:biovid_f1} reports subject-wise F1 scores on the BioVid dataset. Our analysis shows that \ours particularly struggles on subjects where the Physio-only baseline itself is weak (e.g., Sub-4, Sub-5, Sub-7). This suggests that physiological variability across individuals is the key limiting factor: when physio cues are unreliable, visual signals alone are too subtle to compensate, and fusion inherits this weakness. These subject-specific outcomes highlight an inherent challenge of BioVid, where the modality that is usually strongest (physio) can be inconsistent for some individuals, constraining overall adaptation performance. Nevertheless, \ours achieves consistent performance across all subjects and outperforms competing methods on 6 out of 10 targets, leading to the best overall F1 score.

\subsection{BAH}
Tab. \ref{tab:bah_f1} reports subject-wise F1 scores on the BAH dataset. While the visual modality is generally stronger than audio, naive fusion provides slight gains by leveraging complementary information across modalities. \oursmsda achieves the best overall performance, consistently surpassing both lower-bound baselines and existing UDA/MSDA approaches, demonstrating the effectiveness of source subject selection and multimodal adaptation in more challenging, in-the-wild conditions.

\section{Evaluation on Additional Backbone}
In the main paper, all experiments are conducted using a ResNet-18 backbone to ensure fair comparison with prior work and maintain a computationally efficient setup. However, \ours is inherently backbone-agnostic and does not rely on architectural properties specific to convolutional networks. To verify that \ours extends beyond ResNet-based architectures, we additionally evaluate the method using a Vision Transformer (ViT).
We use the same \ours protocol as before for all the experiments, including the same setup for baseline methods. \textbf{Biovid.} Tables~\ref{tab:biovid_vit_acc} and~\ref{tab:biovid_vit_f1} report the subject-wise Accuracy and F1 scores across 10 target subjects. \ours consistently surpasses the baseline UDA and MSDA methods, achieving the highest average Accuracy and F1. Notably, \ours maintains strong performance even in subjects where the baselines degrade, demonstrating stable target adaptation. 
\textbf{StressID.} Tables~\ref{tab:stressid_vit_acc} and~\ref{tab:stressid_vit_f1} present the subject-wise Accuracy and F1 results. \ours consistently outperforms the baseline UDA and MSDA methods across most subjects, achieving the highest average Accuracy and F1. In particular, \ours demonstrates strong improvements on challenging subjects such as Sub-3, Sub-4, and Sub-6, where existing approaches show significant performance drops. These results further validate that the effectiveness of \ours is preserved even when replacing ResNet-18 with a transformer-based backbone, confirming that the proposed method is robust and generalizes well across different architectural families.

\section{\ours}
\ours (inspired by the layered harmony of Musaca) integrates two synergistic modules for subject-specific adaptation: a co-training-based source subject selection module that identifies the most relevant sources using complementary cues from multiple modalities, and an adaptation module that aligns source and target domains using class-aware and class-agnostic losses. Like Musaca’s layered composition—where distinct elements work in harmony, \ours leverages modality-specific strengths to guide pseudo-label generation and fuse information effectively. This enables robust, personalized adaptation for each target subject while maintaining consistency across modalities.

{
    \small
    \bibliographystyle{ieeenat_fullname}
    \bibliography{main}

@String(IJCV = {Int. J. Comput. Vis.})

@String(CVPR= {IEEE Conf. Comput. Vis. Pattern Recog.})

@String(ICPR = {Int. Conf. Pattern Recog.})

@String(IJCV  = {IJCV})

@String(CVPR  = {CVPR})

@String(ICPR  = {ICPR})

@article{rajasekhar2021deep,
  title={Deep domain adaptation with ordinal regression for pain assessment using weakly-labeled videos},
  author={Rajasekhar, Gnana Praveen and Granger, Eric and Cardinal, Patrick},
  journal={Image and Vision Computing},
  volume={110},
  pages={104167},
  year={2021},
  publisher={Elsevier}
}

@inproceedings{li2018deep,
  title={Deep emotion transfer network for cross-database facial expression recognition},
  author={Li, Shan and Deng, Weihong},
  booktitle={24th International Conference on Pattern Recognition (ICPR)},
  pages={3092--3099},
  year={2018},
publisher={IEEE},
address = 	 {Piscataway, USA},
}

@inproceedings{zhu2016discriminative,
  title={Discriminative feature adaptation for cross-domain facial expression recognition},
  author={Zhu, Ronghang and Sang, Gaoli and Zhao, Qijun},
  booktitle={2016 International Conference on Biometrics (ICB)},
  pages={1--7},
  year={2016},
  organization={IEEE}
}

@article{han2020personalized,
  title={Personalized broad learning system for facial expression},
  author={Han, Jing and Xie, Lun and Liu, Jing and Li, Xue},
  journal={Multimedia Tools and Applications},
  volume={79},
  number={23},
  pages={16627--16644},
  year={2020},
  publisher={Springer}
}

@inproceedings{praveen2020deep,
  title={Deep weakly supervised domain adaptation for pain localization in videos},
  author={Praveen, R Gnana and Granger, Eric and Cardinal, Patrick},
  booktitle={2020 15th IEEE International Conference on Automatic Face and Gesture Recognition (FG 2020)},
  pages={473--480},
  year={2020},
  organization={IEEE}
}

@inproceedings{nguyen2021stem,
  title={Stem: An approach to multi-source domain adaptation with guarantees},
  author={Nguyen, Van-Anh and Nguyen, Tuan and Le, Trung and Tran, Quan Hung and Phung, Dinh},
  booktitle={Proceedings of the IEEE/CVF International Conference on Computer Vision},
  pages={9352--9363},
  year={2021},
publisher={IEEE},
address = 	 {Piscataway, USA},
}

@article{zhao2021madan,
  title={MADAN: multi-source adversarial domain aggregation network for domain adaptation},
  author={Zhao, Sicheng and Li, Bo and Xu, Pengfei and Yue, Xiangyu and Ding, Guiguang and Keutzer, Kurt},
  journal={IJCV},
  volume={129},
  number={8},
  pages={2399--2424},
  year={2021},
  publisher={Springer}
}

@inproceedings{walter2013biovid,
  title={The biovid heat pain database data for the advancement and systematic validation of an automated pain recognition system},
  author={Walter, Steffen and Gruss, Sascha and Ehleiter, Hagen and Tan, Junwen and Traue, Harald C and Werner, Philipp and Al-Hamadi, Ayoub and Crawcour, Stephen and Andrade, Adriano O and da Silva, Gustavo Moreira},
  booktitle={2013 IEEE international conference on cybernetics (CYBCO)},
  pages={128--131},
  year={2013},
  organization={IEEE}
}

@inproceedings{peng2019moment,
  title={Moment matching for multi-source domain adaptation},
  author={Peng, Xingchao and Bai, Qinxun and Xia, Xide and Huang, Zijun and Saenko, Kate and Wang, Bo},
  booktitle={Proceedings of the IEEE/CVF international conference on computer vision},
  pages={1406--1415},
  year={2019},
publisher={IEEE},
address = 	 {Piscataway, USA},
}

@article{ren2022multi,
  title={Multi-source unsupervised domain adaptation via pseudo target domain},
  author={Ren, Chuan-Xian and Liu, Yong-Hui and Zhang, Xi-Wen and Huang, Ke-Kun},
  journal={IEEE Transactions on Image Processing},
  volume={31},
  pages={2122--2135},
  year={2022},
  publisher={IEEE}
}

@article{deng2022robust,
  title={Robust Target Training for Multi-Source Domain Adaptation},
  author={Deng, Zhongying and Li, Da and Song, Yi-Zhe and Xiang, Tao},
  journal={CoRR},
volumne={abs/:2210.01676},
  year={2022}
}

@article{kang2020contrastive,
  title={Contrastive adaptation network for single-and multi-source domain adaptation},
  author={Kang, Guoliang and Jiang, Lu and Wei, Yunchao and Yang, Yi and Hauptmann, Alexander},
  journal={IEEE TPAMI},
  volume={44},
  number={4},
  pages={1793--1804},
  year={2020},
  publisher={IEEE}
}

@article{venkat2020your,
  title={Your classifier can secretly suffice multi-source domain adaptation},
  author={Venkat, Naveen and Kundu, Jogendra Nath and Singh, Durgesh and Revanur, Ambareesh and others},
  journal={Advances in Neural Information Processing Systems},
  volume={33},
  pages={4647--4659},
  year={2020}
}

@article{scalbert2021multi,
  title={Multi-source domain adaptation via supervised contrastive learning and confident consistency regularization},
  author={Scalbert, Marin and Vakalopoulou, Maria and Couzini{\'e}-Devy, Florent},
  journal={arXiv preprint arXiv:2106.16093},
  year={2021}
}

@inproceedings{he2016deep,
  title={Deep residual learning for image recognition},
  author={He, Kaiming and Zhang, Xiangyu and Ren, Shaoqing and Sun, Jian},
  booktitle={Proceedings of the IEEE conference on computer vision and pattern recognition},
  pages={770--778},
  year={2016},
publisher={IEEE},
address = 	 {Piscataway, USA},
}

@article{gretton2012kernel,
  title={A kernel two-sample test},
  author={Gretton, Arthur and Borgwardt, Karsten M and Rasch, Malte J and Sch{\"o}lkopf, Bernhard and Smola, Alexander},
  journal={The Journal of Machine Learning Research},
  volume={13},
  number={1},
  pages={723--773},
  year={2012},
  publisher={JMLR. org}
}

@article{long2016unsupervised,
  title={Unsupervised domain adaptation with residual transfer networks},
  author={Long, Mingsheng and Zhu, Han and Wang, Jianmin and Jordan, Michael I},
  journal={Advances in neural information processing systems},
  volume={29},
  year={2016}
}

@inproceedings{xu2018deep,
  title={Deep cocktail network: Multi-source unsupervised domain adaptation with category shift},
  author={Xu, Ruijia and Chen, Ziliang and Zuo, Wangmeng and Yan, Junjie and Lin, Liang},
  booktitle={Proceedings of the IEEE conference on computer vision and pattern recognition},
  pages={3964--3973},
  year={2018}
}

@article{sejdinovic2013equivalence,
  title={Equivalence of distance-based and RKHS-based statistics in hypothesis testing},
  author={Sejdinovic, Dino and Sriperumbudur, Bharath and Gretton, Arthur and Fukumizu, Kenji},
  journal={The annals of statistics},
  pages={2263--2291},
  year={2013},
  publisher={JSTOR}
}

@inproceedings{werner2017analysis,
  title={Analysis of facial expressiveness during experimentally induced heat pain},
  author={Werner, Philipp and Al-Hamadi, Ayoub and Walter, Steffen},
  booktitle={2017 Seventh international conference on affective computing and intelligent interaction workshops and demos (ACIIW)},
  pages={176--180},
  year={2017},
  organization={IEEE}
}

@article{van2008visualizing,
  title={Visualizing data using t-SNE.},
  author={Van der Maaten, Laurens and Hinton, Geoffrey},
  journal={Journal of machine learning research},
  volume={9},
  number={11},
  year={2008}
}

@inproceedings{zeeshan2024subject,
  title={Subject-based domain adaptation for facial expression recognition},
  author={Zeeshan, Muhammad Osama and Aslam, Muhammad Haseeb and Belharbi, Soufiane and Koerich, Alessandro Lameiras and Pedersoli, Marco and Bacon, Simon and Granger, Eric},
  booktitle={2024 IEEE 18th International Conference on Automatic Face and Gesture Recognition (FG)},
  pages={1--10},
  year={2024},
  organization={IEEE}
}

@inproceedings{aslam2024distilling,
  title={Distilling privileged multimodal information for expression recognition using optimal transport},
  author={Aslam, Muhammad Haseeb and Zeeshan, Muhammad Osama and Belharbi, Soufiane and Pedersoli, Marco and Koerich, Alessandro Lameiras and Bacon, Simon and Granger, Eric},
  booktitle={18th International Conference on Automatic Face and Gesture Recognition (FG)},
  pages={1--10},
  year={2024}
}

@inproceedings{aslam2023privileged,
  title={Privileged knowledge distillation for dimensional emotion recognition in the wild},
  author={Aslam, Muhammad Haseeb and Zeeshan, Muhammad Osama and Pedersoli, Marco and Koerich, Alessandro L and Bacon, Simon and Granger, Eric},
  booktitle={Proceedings of the IEEE/CVF conference on computer vision and pattern recognition},
  pages={3338--3347},
  year={2023}
}

@inproceedings{waligora2024joint,
  title={Joint Multimodal Transformer for Emotion Recognition in the Wild},
  author={Waligora, Paul and Aslam, Muhammad Haseeb and Zeeshan, Muhammad Osama and Belharbi, Soufiane and Koerich, Alessandro Lameiras and Pedersoli, Marco and Bacon, Simon and Granger, Eric},
  booktitle={Proceedings of the IEEE/CVF Conference on Computer Vision and Pattern Recognition},
  pages={4625--4635},
  year={2024}
}

@article{liu2021comparing,
  title={Comparing recognition performance and robustness of multimodal deep learning models for multimodal emotion recognition},
  author={Liu, Wei and Qiu, Jie-Lin and Zheng, Wei-Long and Lu, Bao-Liang},
  journal={IEEE Transactions on Cognitive and Developmental Systems},
  volume={14},
  number={2},
  pages={715--729},
  year={2021},
  publisher={IEEE}
}

@article{zhang2020emotion,
  title={Emotion recognition using multi-modal data and machine learning techniques: A tutorial and review},
  author={Zhang, Jianhua and Yin, Zhong and Chen, Peng and Nichele, Stefano},
  journal={Information Fusion},
  volume={59},
  pages={103--126},
  year={2020},
  publisher={Elsevier}
}

@article{dzedzickis2020human,
  title={Human emotion recognition: Review of sensors and methods},
  author={Dzedzickis, Andrius and Kaklauskas, Art{\=u}ras and Bucinskas, Vytautas},
  journal={Sensors},
  volume={20},
  number={3},
  pages={592},
  year={2020},
  publisher={MDPI}
}

@article{tzirakis2017end,
  title={End-to-end multimodal emotion recognition using deep neural networks},
  author={Tzirakis, Panagiotis and Trigeorgis, George and Nicolaou, Mihalis A and Schuller, Bj{\"o}rn W and Zafeiriou, Stefanos},
  journal={Journal of selected topics in signal processing},
  volume={11},
  number={8},
  pages={1301--1309},
  year={2017},
  publisher={IEEE}
}

@inproceedings{praveen2022joint,
  title={A joint cross-attention model for audio-visual fusion in dimensional emotion recognition},
  author={Praveen, R Gnana and de Melo, Wheidima Carneiro and Ullah, Nasib and Aslam, Haseeb and Zeeshan, Osama and Denorme, Th{\'e}o and Pedersoli, Marco and Koerich, Alessandro L and Bacon, Simon and Cardinal, Patrick and others},
  booktitle={Proceedings of the IEEE/CVF CVPR},
  pages={2486--2495},
  year={2022}
}

@inproceedings{pichler2022differential,
  title={A differential entropy estimator for training neural networks},
  author={Pichler, Georg and Colombo, Pierre Jean A and Boudiaf, Malik and Koliander, G{\"u}nther and Piantanida, Pablo},
  booktitle={International Conference on Machine Learning},
  pages={17691--17715},
  year={2022},
  organization={PMLR},
publisher={PMLR}

}

@article{chaptoukaev2023stressid,
  title={Stressid: a multimodal dataset for stress identification},
  author={Chaptoukaev, Hava and Strizhkova, Valeriya and Panariello, Michele and Dalpaos, Bianca and Reka, Aglind and Manera, Valeria and Th{\"u}mmler, Susanne and Ismailova, Esma and Todisco, Massimiliano and Zuluaga, Maria A and others},
  journal={Advances in Neural Information Processing Systems},
  volume={36},
  pages={29798--29811},
  year={2023}
}

@article{jiang2022disentangling,
  title={Disentangling identity and pose for facial expression recognition},
  author={Jiang, Jing and Deng, Weihong},
  journal={IEEE Transactions on Affective Computing},
  volume={13},
  number={4},
  pages={1868--1878},
  year={2022},
  publisher={IEEE}
}

@article{xie2020facial,
  title={Facial expression recognition with two-branch disentangled generative adversarial network},
  author={Xie, Siyue and Hu, Haifeng and Chen, Yizhen},
  journal={IEEE Transactions on Circuits and Systems for Video Technology},
  volume={31},
  number={6},
  pages={2359--2371},
  year={2020},
  publisher={IEEE}
}

@article{wang2022multi,
  title={Multi-modal domain adaptation variational autoencoder for EEG-based emotion recognition},
  author={Wang, Yixin and Qiu, Shuang and Li, Dan and Du, Changde and Lu, Bao-Liang and He, Huiguang},
  journal={IEEE/CAA Journal of Automatica Sinica},
  volume={9},
  number={9},
  pages={1612--1626},
  year={2022},
  publisher={IEEE}
}

@article{jimenez2024cfda,
  title={CFDA-CSF: A multi-modal domain adaptation method for cross-subject emotion recognition},
  author={Jim{\'e}nez-Guarneros, Magdiel and Fuentes-Pineda, Gibran},
  journal={IEEE Transactions on Affective Computing},
  volume={15},
  number={3},
  pages={1502--1513},
  year={2024},
  publisher={IEEE}
}

@article{chen2011co,
  title={Co-training for domain adaptation},
  author={Chen, Minmin and Weinberger, Kilian Q and Blitzer, John},
  journal={Advances in neural information processing systems},
  volume={24},
  year={2011}
}

@inproceedings{blum1998combining,
  title={Combining labeled and unlabeled data with co-training},
  author={Blum, Avrim and Mitchell, Tom},
  booktitle={Proceedings of the eleventh annual conference on Computational learning theory},
  pages={92--100},
  year={1998}
}

@article{jimenez2024mmda,
  title={MMDA: A Multimodal and Multisource Domain Adaptation Method for Cross-Subject Emotion Recognition From EEG and Eye Movement Signals},
  author={Jim{\'e}nez-Guarneros, Magdiel and Fuentes-Pineda, Gibran and Grande-Barreto, Jonas},
  journal={IEEE Transactions on Computational Social Systems},
  year={2024},
  publisher={IEEE}
}

@article{gomez2023co,
  title={Co-training for unsupervised domain adaptation of semantic segmentation models},
  author={G{\'o}mez, Jose L and Villalonga, Gabriel and L{\'o}pez, Antonio M},
  journal={Sensors},
  volume={23},
  number={2},
  pages={621},
  year={2023},
  publisher={MDPI}
}

@ARTICLE{zeeshan2025progressive,
  author={Zeeshan, Muhammad Osama and Pedersoli, Marco and Lameiras Koerich, Alessandro and Granger, Eric},
  journal={IEEE Transactions on Affective Computing}, 
  title={Progressive Multi-Source Domain Adaptation for Personalized Facial Expression Recognition}, 
  year={2025},
  doi={10.1109/TAFFC.2025.3627534}}

@inproceedings{yang2021deep,
  title={Deep co-training with task decomposition for semi-supervised domain adaptation},
  author={Yang, Luyu and Wang, Yan and Gao, Mingfei and Shrivastava, Abhinav and Weinberger, Kilian Q and Chao, Wei-Lun and Lim, Ser-Nam},
  booktitle={Proceedings of the IEEE/CVF international conference on computer vision},
  pages={8906--8916},
  year={2021}
}

@article{xia2020uncertainty,
  title={Uncertainty-aware multi-view co-training for semi-supervised medical image segmentation and domain adaptation},
  author={Xia, Yingda and Yang, Dong and Yu, Zhiding and Liu, Fengze and Cai, Jinzheng and Yu, Lequan and Zhu, Zhuotun and Xu, Daguang and Yuille, Alan and Roth, Holger},
  journal={Medical image analysis},
  volume={65},
  pages={101766},
  year={2020},
  publisher={Elsevier}
}

@inproceedings{sharafi2025disentangled,
  title={Disentangled source-free personalization for facial expression recognition with neutral target data},
  author={Sharafi, Masoumeh and Ollivier, Emma and Zeeshan, Muhammad Osama and Belharbi, Soufiane and Koerich, Alessandro Lameiras and Pedersoli, Marco and Bacon, Simon and Granger, Eric},
  booktitle={2025 IEEE 19th International Conference on Automatic Face and Gesture Recognition (FG)},
  pages={1--10},
  year={2025},
  organization={IEEE}
}

@article{ruder2016overview,
  title={An overview of gradient descent optimization algorithms},
  author={Ruder, Sebastian},
  journal={arXiv preprint arXiv:1609.04747},
  year={2016}
}

@article{ganin2016domain,
  title={Domain-adversarial training of neural networks},
  author={Ganin, Yaroslav and Ustinova, Evgeniya and Ajakan, Hana and Germain, Pascal and Larochelle, Hugo and Laviolette, Fran{\c{c}}ois and March, Mario and Lempitsky, Victor},
  journal={Journal of machine learning research},
  volume={17},
  number={59},
  pages={1--35},
  year={2016}
}

@article{long2018conditional,
  title={Conditional adversarial domain adaptation},
  author={Long, Mingsheng and Cao, Zhangjie and Wang, Jianmin and Jordan, Michael I},
  journal={NeurIPS},
  volume={31},
  year={2018}
}

@article{jack2009cultural,
  title={Cultural confusions show that facial expressions are not universal},
  author={Jack, Rachael E and Blais, Caroline and Scheepers, Christoph and Schyns, Philippe G and Caldara, Roberto},
  journal={Current biology},
  volume={19},
  number={18},
  pages={1543--1548},
  year={2009},
  publisher={Elsevier}
}

@article{gonzalez2025bah,
  title={BAH Dataset for Ambivalence/Hesitancy Recognition in Videos for Behavioural Change},
  author={Gonz{\'a}lez-Gonz{\'a}lez, Manuela and Belharbi, Soufiane and Zeeshan, Muhammad Osama and Sharafi, Masoumeh and Aslam, Muhammad Haseeb and Pedersoli, Marco and Koerich, Alessandro Lameiras and Bacon, Simon L and Granger, Eric},
  journal={arXiv preprint arXiv:2505.19328},
  year={2025}
}

@article{kong2020panns,
  title={Panns: Large-scale pretrained audio neural networks for audio pattern recognition},
  author={Kong, Qiuqiang and Cao, Yin and Iqbal, Turab and Wang, Yuxuan and Wang, Wenwu and Plumbley, Mark D},
  journal={IEEE/ACM Transactions on Audio, Speech, and Language Processing},
  volume={28},
  pages={2880--2894},
  year={2020},
  publisher={IEEE}
}

@inproceedings{guichemerre2024source,
  title={Source-free domain adaptation of weakly-supervised object localization models for histology},
  author={Guichemerre, Alexis and Belharbi, Soufiane and Mayet, Tsiry and Murtaza, Shakeeb and Shamsolmoali, Pourya and McCaffrey, Luke and Granger, Eric},
  booktitle={Proceedings of the IEEE/CVF Conference on Computer Vision and Pattern Recognition},
  pages={33--43},
  year={2024}
}
}
\end{document}